\documentclass{article}

    \PassOptionsToPackage{numbers, compress}{natbib}

\usepackage[preprint]{neurips_2025_rs}

\newcommand{\degC}{\ensuremath{^\circ\mathrm{C}~}}

\usepackage[table,xcdraw]{xcolor}

\definecolor{darkcerulean}{rgb}{0.03, 0.27, 0.49}
\definecolor{smokyblack}{rgb}{0.06, 0.05, 0.03}
\definecolor{warmblack}{rgb}{0.0, 0.26, 0.26}
\definecolor{cobalt}{rgb}{0.0, 0.28, 0.67}
\definecolor{darkerred}{RGB}{139, 0, 0}
\definecolor{darkergreen}{RGB}{0, 100, 0}
\definecolor{warmred}{RGB}{205, 92, 92}
\definecolor{warmgreen}{RGB}{34, 139, 34}
\definecolor{lightyellow}{RGB}{255, 255, 224}
\definecolor{warmyellow}{RGB}{255, 250, 205}
\definecolor{flax}{RGB}{238, 220, 130}
\definecolor{warmblue}{RGB}{70, 130, 180}
\definecolor{arrowredfig1}{HTML}{B85450}
\definecolor{expertfig1}{HTML}{6A9153}
\definecolor{mixturefig1}{HTML}{6C8EBF}

\newcommand{\colorwblk}[1]{\textcolor{warmblack}{#1}}
\newcommand{\colorwr}[1]{\textcolor{warmred}{#1}}
\newcommand{\colorwg}[1]{\textcolor{warmgreen}{#1}}

\newcommand{\colorwb}[1]{\textcolor{warmblue}{#1}}

\newcommand{\colorwgb}[1]{\textcolor{warmgreen}{\textbf{#1}}}
\newcommand{\colorwblkb}[1]{\textcolor{warmblack}{\textbf{#1}}}

\newcommand{\cmark}{\textcolor{green!60!black}{\ding{51}}}
\newcommand{\xmark}{\textcolor{red!75!black}{\ding{55}}}

\usepackage[colorinlistoftodos]{todonotes}

\usepackage{ifthen}
\usepackage{tcolorbox}
\tcbuselibrary{breakable}
\usepackage{lettrine}

\usepackage[T1]{fontenc}
\usepackage[utf8]{inputenc}
\usepackage{fontawesome}
\usepackage{pifont}

\usepackage{amsfonts}
\usepackage{amsmath}
\usepackage{amssymb}
\usepackage[]{bbm}

\usepackage{booktabs}
\usepackage{graphicx}
\usepackage[format=default,font=small,labelfont=it]{caption}

\usepackage{subcaption}
\usepackage{makecell}
\usepackage{placeins}

\usepackage{pdflscape}
\usepackage{siunitx}
\usepackage[bottom]{footmisc}
\usepackage[normalem]{ulem}
\usepackage[toc,page]{appendix}
\usepackage{array}
\usepackage{blindtext}
\usepackage{bm}
\usepackage{enumitem}
\usepackage{epigraph}
\usepackage{lipsum}
\usepackage{mathtools}
\usepackage{microtype}
\usepackage{nicefrac}
\usepackage{nicematrix,tikz}
\usepackage{setspace}
\usepackage{sidecap}
\usepackage{url}
\usepackage{wrapfig}

\usepackage{adjustbox}
\usepackage[UKenglish]{isodate}
\usepackage{silence}

\usepackage{fancyvrb}
\usepackage{rotating}

\usepackage{titletoc}
\newcommand\DoToC{
  \startcontents
  \printcontents{}{1}{\textbf{Appendix Contents}\vskip3pt\hrule\vskip5pt}
  \vskip3pt\hrule\vskip5pt
}

\usepackage{listings}
\lstnewenvironment{rsverbatim}[1][]{
  \lstset{
    basicstyle=\small\ttfamily,
    columns=flexible,
    breaklines=true,
    frame=tb,
    #1
  }
}{}

\definecolor{MidnightBlue}{RGB}{25,25,112}
\definecolor{MidnightBlueComplementingGreen}{RGB}{25,112,25}
\definecolor{MidnightBlueComplementingPurple}{RGB}{112,25,112}
\definecolor{amaranth}{rgb}{0.9, 0.17, 0.31}
\definecolor{MidnightBlueComplementingRed}{RGB}{112,25,69}
\definecolor{coolblack}{rgb}{0.0, 0.18, 0.39}

\definecolor{deepjunglegreen}{rgb}{0.0, 0.29, 0.29}
\definecolor{applegreen}{rgb}{0.55, 0.71, 0.0}
\definecolor{WowColor}{rgb}{.75,0,.75}
\definecolor{MildlyAlarming}{rgb}{0.85,0.25,0.1}
\definecolor{SubtleColor}{rgb}{0,0,.50}
\definecolor{SubtleColor2}{rgb}{0.6,0.21,.50}

\definecolor{lasallegreen}{rgb}{0.03, 0.47, 0.19}

\newcounter{margincounter}

\definecolor{CBdarkgreen}{rgb}{0.11, 0.54, 0.18}

\NewDocumentCommand{\AK}{mo}{
    \IfValueF{#2}{

        {{\scriptsize
            \textcolor{violet}{
            \textbf{A:}
            \textit{{#1}}
            }
        }}
    }

    \IfValueT{#2}{
        \marginnote{{\scriptsize
            \textcolor{violet}{
            \textbf{A:}
            \textit{{#1}}
            }
        }}
    }
}

\definecolor{darkgreen}{rgb}{0.0, 0.2, 0.13}

\usepackage[linesnumbered,lined,boxed,commentsnumbered,ruled,longend]{algorithm2e}

\usepackage{tikz}
\usetikzlibrary{matrix,chains,positioning,arrows, calc}

\usepackage{float}
\usepackage{amsthm}
\usepackage{multirow}
\usepackage{multicol}

\usepackage[framemethod=TikZ]{mdframed}

\newcounter{defn}[section] 
\renewcommand{\thedefn}{\arabic{section}.\arabic{defn}}

\newcounter{theo}[section] 
\renewcommand{\thetheo}{\arabic{section}.\arabic{theo}}

\newcounter{lem}[section] 

\renewcommand{\thelem}{\arabic{lem}}

\newcounter{prf}[section]
\renewcommand{\theprf}{\arabic{section}.\arabic{prf}}

\theoremstyle{remark}

\theoremstyle{definition}

\newcommand{\BlackBox}{\rule{1.5ex}{1.5ex}}
\ifdefined\proof
    \renewenvironment{proof}{\par\noindent{\bf Proof\ }}{\hfill\BlackBox\\[2mm]}
\else
    \newenvironment{proof}{\par\noindent{\bf Proof\ }}{\hfill\BlackBox\\[2mm]}
\fi

\usepackage[colorlinks=true,
                linkcolor=cobalt,
                urlcolor=darkcerulean,
                citecolor=warmblack]{hyperref}

\usepackage{float}
\usepackage{enumitem}
\usepackage{multirow}

    \usepackage{tikz}
    \usepackage{tikz-cd}
    \pgfdeclarelayer{edgelayer}
    \pgfdeclarelayer{nodelayer}
    \pgfsetlayers{edgelayer,nodelayer,main}

    \tikzstyle{new style 0}=[fill={rgb,255: red,255; green,94; blue,247}, draw=black, shape=circle]
    \tikzstyle{pointy}=[fill=white, draw=black, shape=circle]

    \tikzstyle{pointy}=[->]
\usepackage{fontawesome}
\usepackage{appendix}
\usepackage{cleveref}
\usepackage{comment}

\usepackage{algpseudocode}
\usepackage[linesnumbered,lined,boxed,commentsnumbered,ruled,longend]{algorithm2e}

\SetCommentSty{mycommfont}

\newcommand{\pushright}[1]{\ifmeasuring@#1\else\omit\hfill$\displaystyle#1$\fi\ignorespaces}
\newcommand{\pushleft}[1]{\ifmeasuring@#1\else\omit$\displaystyle#1$\hfill\fi\ignorespaces}
\makeatother

\newcommand{\1}{\mathbbm{1}}

\renewcommand{\phi}{\varphi}

\RequirePackage{amsthm}

\theoremstyle{remark}

\NewDocumentCommand{\luca}{mo}{
    \IfValueF{#2}{

                        {{\scriptsize
                            \textcolor{green}{
                            \textbf{L:}
                            \textit{{#1}}
                            }
                        }}
        }

    \IfValueT{#2}{
                        \marginnote{{\scriptsize
                            \textcolor{green}{
                            \textbf{L:}
                            \textit{{#1}}
                            }
                        }}
        }
                    }
\NewDocumentCommand{\giulia}{mo}{
    \IfValueF{#2}{

                        {{\scriptsize
                            \textcolor{red}{
                            \textbf{GL:}
                            \textit{{#1}}
                            }
                        }}
        }

    \IfValueT{#2}{
                        \marginnote{{\scriptsize
                            \textcolor{red}{
                            \textbf{GL:}
                            \textit{{#1}}
                            }
                        }}
        }
}
\NewDocumentCommand{\anastasis}{mo}{
    \IfValueF{#2}{

                        {{\scriptsize
                            \textcolor{violet}{
                            \textbf{A:}
                            \textit{{#1}}
                            }
                        }}
        }

    \IfValueT{#2}{
                        \marginnote{{\scriptsize
                            \textcolor{violet}{
                            \textbf{A:}
                            \textit{{#1}}
                            }
                        }}
        }
                    }

\NewDocumentCommand{\cody}{mo}{
    \IfValueF{#2}{

                        {{\scriptsize
                            \textcolor{orange}{
                            \textbf{A:}
                            \textit{{#1}}
                            }
                        }}
        }

    \IfValueT{#2}{
                        \marginnote{{\scriptsize
                            \textcolor{orange}{
                            \textbf{A:}
                            \textit{{#1}}
                            }
                        }}
        }
                    }

\NewDocumentCommand{\yannick}{mo}{
    \IfValueF{#2}{

                        {{\scriptsize
                            \textcolor{cyan}{
                            \textbf{Y:}
                            \textit{{#1}}
                            }
                        }}
        }

    \IfValueT{#2}{
                        \marginnote{{\scriptsize
                            \textcolor{cyan}{
                            \textbf{Y:}
                            \textit{{#1}}
                            }
                        }}
        }
                    }

\definecolor{darkgreen}{rgb}{0.0, 0.2, 0.13}
\NewDocumentCommand{\xuwei}{mo}{
    \IfValueF{#2}{

                        {{\scriptsize
                            \textcolor{darkgreen}{
                            \textbf{X:}
                            \textit{{#1}}
                            }
                        }}
        }

    \IfValueT{#2}{
                        \marginnote{{\scriptsize
                            \textcolor{darkgreen}{
                            \textbf{X:}
                            \textit{{#1}}
                            }
                        }}
        }
                    }

\newcounter{termcounter}
\renewcommand{\thetermcounter}{\Roman{termcounter}}
\crefname{term}{term}{terms}
\creflabelformat{term}{#2\textup{(#1)}#3}

\makeatletter
\def\term{\@ifnextchar[\term@optarg\term@noarg}
\def\term@optarg[#1]#2{
  \textup{#1}
  \def\@currentlabel{#1}
  \def\cref@currentlabel{[][2147483647][]#1}
  \cref@label[term]{#2}}
\def\term@noarg#1{
  \refstepcounter{termcounter}
  \textup{(\thetermcounter)}
  \cref@label[term]{#1}}
\makeatother

\crefname{lemma}{lemma}{lemmata}
\Crefname{lemma}{Lemma}{Lemmata}
\crefname{assumption}{assumption}{assumptions}
\Crefname{assumption}{Assumption}{Assumptions}
\crefname{example}{Example}{Examples}
\crefname{proposition}{Proposition}{Proposition}

\def\eqref#1{equation~\ref{#1}}

\def\1{\bm{1}}

\DeclareMathAlphabet{\mathsfit}{\encodingdefault}{\sfdefault}{m}{sl}
\SetMathAlphabet{\mathsfit}{bold}{\encodingdefault}{\sfdefault}{bx}{n}

\graphicspath{ {figs/} }

\setboolean{use_header_icon}{true}

\ifthenelse{\boolean{use_header_icon}}{
  \papertitlewithimage{figs/hamster_seeking-sota2_sm}{Seeking SOTA: Time-Series Forecasting Must Adopt Taxonomy-Specific Evaluation to Dispel Illusory Gains.}
}{
  \title{Seeking SOTA: Time-Series Forecasting Must Adopt Taxonomy-Specific Evaluation to Dispel Illusory Gains.}
}

\author{
  Raeid Saqur\textsuperscript{1,4,6}\thanks{Corresponding authors: \href{mailto:raeidsaqur@cs.toronto.edu}{\texttt{raeidsaqur@cs.toronto.edu}}, \href{mailto:bergmeir@ugr.es}{\texttt{bergmeir@ugr.es}}} \quad
  Christoph Bergmeir\textsuperscript{2,7,8}\quad
  \And
  Blanka Horvath\textsuperscript{4,5} \quad
  Daniel Schmidt\textsuperscript{2} \quad
  Frank Rudzicz\textsuperscript{3,6} \quad
  Terry Lyons\textsuperscript{4} \\
  \vspace{1pt}\\
  \small \textsuperscript{1}Dept. of Computer Science, University of Toronto \\
  \small \textsuperscript{2}Dept. of Data Science \& AI, Monash University, Australia\\
  \small \textsuperscript{3}Faculty of Computer Science, Dalhousie University \\
  \small \textsuperscript{4}Dept. of Mathematics, University of Oxford \\
  \small \textsuperscript{5}Oxford-Man Institute for Quantitative Finance \\
  \small \textsuperscript{6}Vector Institute \\
  \small \textsuperscript{7}Department of Computer Science and AI, University of Granada, Spain.\\
  \small \textsuperscript{8}DaSCI, Andalucía, Spain. \\
  \normalsize
}

\begin{document}
\maketitle

\begin{abstract}
We argue that the current practice of evaluating AI/ML time-series forecasting models --- predominantly on benchmarks characterized by strong, persistent periodicities and seasonalities --- obscures real progress by overlooking the performance of efficient, classical methods. We demonstrate that these ``standard'' datasets often exhibit dominant auto-correlation patterns and seasonal cycles that can be effectively captured by simpler linear or statistical models, rendering complex deep-learning architectures frequently no more performant than their classical counterparts for these specific data characteristics, \colorwblkb{and critically, raising questions as to whether any marginal improvements justify the significant increase in computational overhead and model complexity.} We call on the community to \colorwblkb{(I)} retire or substantially augment current benchmarks with datasets exhibiting a wider spectrum of non-stationarities (e.g., structural breaks, time-varying volatility, concept drift) and less predictable dynamics drawn from diverse real-world domains, and \colorwblkb{(II)} require every deep-learning submission to include robust classical and simple baselines, appropriately chosen for the specific characteristics (e.g., type of non-stationarity, seasonality) of the downstream tasks' time series. By doing so, we will ensure that reported gains reflect genuine scientific methodological advances rather than artifacts of benchmark selection favoring models adept at learning repetitive patterns.
\end{abstract}

\section{Introduction}\label{sec:intro}

\lettrine{G}{oodhart’s} Law warns us: \textit{when a measure becomes a target, it ceases to be a good measure}. In scientific research, this becomes increasingly evident as AI/ML fields rapidly expand, with optimization on benchmark metrics often supplanting broader notions of meaningful progress. When benchmarks are static, narrow, or unrepresentative of real-world complexity, they risk distorting rather than reflecting genuine innovation. This paper highlights the insidious nature of problematic evaluation schemes prevalent in recent long-horizon time series forecasting (LTSF) research and takes a firm stand: ``\colorwblkb{Time-Series Forecasting Must Adopt Taxonomy-Specific Evaluation to Dispel Illusory Gains from Periodicity-Dominant Benchmarks}''.

Over the past five years, a significant number of papers in top-tier AI/ML conferences have proposed increasingly elaborate neural architectures for time-series forecasting (TSF). These include transformer-based models~\cite{tsf_zhou2021informer, tsf_wu2021autoformer, tsf_liu2021pyraformer, tsf_zhou2022fedformer, tsf_sparsetsf-lin24n} and Large Language Model (LLM) \textit{aka} Foundation Model~(FM)~\cite{bommasani2021opportunities} inspired TSF approaches~\cite{tsf_timellm_iclr24_jin2023time, tsf_aaai25_liu2025calf}. Despite a handful of recent critiques~\cite{tsf_critique_tan2024language, tsf_critique_zeng2023transformers} exposing issues with purported progress in TSF using these methods—transformer-variants and LLMs—a steady surge of subsequent peer-reviewed works continues to uncritically build upon them. It is therefore imperative for the community to clearly understand these issues and halt the further accumulation of \colorwblk{\textbf{\textit{cascading errors}}}.

For our pointed arguments and illustration, we utilize the nine so-called standard long-horizon TSF (LTSF) benchmark datasets detailed in Table~\ref{tab:tsf-dataset_statistics}. Introduced circa 2021~\cite{tsf_zhou2021informer}, these datasets rapidly became the \textit{de facto} standard for evaluating subsequent LTSF models. Table~\ref{tab:complexity_comparison_transformer_tsfs} provides historical context by listing several highly-cited early TSF transformer-variants (circa 2021-2022) that adopted and popularized this benchmark, along with their attention module complexities.

\begin{center}
\begin{minipage}[t]{0.50\linewidth}
    \centering
    \renewcommand{\arraystretch}{1.2}
    \captionof{table}{Statistics of Nine Multivariate \colorwb{LTSF Datasets}.}
    \label{tab:tsf-dataset_statistics}
    \resizebox{\linewidth}{!}{
        \begin{tabular}{@{}lccc@{}}
            \toprule
            \textbf{Dataset(s)} & \textbf{Channels} & \textbf{Timesteps} & \textbf{Sampling-Rate} \\
            \midrule
            ETTh1 \& ETTh2~\cite{tsf_zhou2021informer} & 7 & 17,420 & 1 hour \\
            ETTm1 \& ETTm2~\cite{tsf_zhou2021informer} & 7 & 69,680 & 5 min \\
            Traffic~\cite{tsf_dataset_traffic} & 862 & 17,544 & 1 hour \\
            Electricity~\cite{tsf_dataset_electricity} & 321 & 26,304 & 1 hour \\
            Exchange-Rate~\cite{tsf_dataset_exchange_lai2018modeling} & 8 & 7,588 & 1 day \\
            Weather~\cite{tsf_dataset_weather_angryk2020multivariate} & 21 & 52,696 & 10 min \\
            Illness (ILI)~\cite{tsf_dataset_ili_poorawala2022novel} & 7 & 966 & 1 week \\
            \bottomrule
        \end{tabular}
    }
\end{minipage}
\hfill
\begin{minipage}[t]{0.45\linewidth}
    \centering
    \renewcommand{\arraystretch}{1.2}
    \captionof{table}{Early TSF Transformer Complexities.}
    \label{tab:complexity_comparison_transformer_tsfs}
    \resizebox{\linewidth}{!}{
        \begin{tabular}{@{}lccc@{}}
            \toprule
            \textbf{Method} & \multicolumn{2}{c}{Training Complexity} & \textbf{Test Steps} \\
            \cmidrule(r){2-3}
                     & Time & Memory &  \\
            \midrule
            Transformer~\cite{vaswani2017attention} & $\mathcal{O}(N^2)$ & $\mathcal{O}(N^2)$ & $N$ \\
            LogTrans~\cite{tsf_logsparsetransformer_li2019enhancing} & $\mathcal{O}(N \log N)$ & $\mathcal{O}(N (\log N)^2)$ & $1$ \\
            Informer~\cite{tsf_zhou2021informer} & $\mathcal{O}(N \log N)$ & $\mathcal{O}(N \log N)$ & $1$ \\
            Autoformer~\cite{tsf_wu2021autoformer} & $\mathcal{O}(N \log N)$ & $\mathcal{O}(N \log N)$ & $1$ \\
            Pyraformer~\cite{tsf_liu2021pyraformer} & $\mathcal{O}(N)$ & $\mathcal{O}(N)$ & $1$ \\
            FEDformer~\cite{tsf_zhou2022fedformer} & $\mathcal{O}(N)$ & $\mathcal{O}(N)$ & $1$ \\
            \bottomrule
        \end{tabular}
    }
\end{minipage}
\end{center}
\vspace{1em}

\begin{figure*}[t!]
    \centering
    \setlength{\tabcolsep}{1pt}
    \renewcommand{\arraystretch}{0.5}
    \begin{tabular}{@{}c@{\hspace{2pt}}c@{\hspace{2pt}}c@{\hspace{2pt}}c@{\hspace{2pt}}c@{}}
        \multicolumn{1}{c}{\small\textbf{\textsc{ETT}h2}} &
        \multicolumn{1}{c}{\small\textbf{\textsc{ETT}m2}} &
        \multicolumn{1}{c}{\small\textbf{\textsc{Traffic}}} &
        \multicolumn{1}{c}{\small\textbf{\textsc{Exchange}}} &
        \multicolumn{1}{c}{\small\textbf{\textsc{ILI}}} \\

        \begin{subfigure}[b]{0.19\textwidth}
            \centering
            \includegraphics[width=\linewidth]{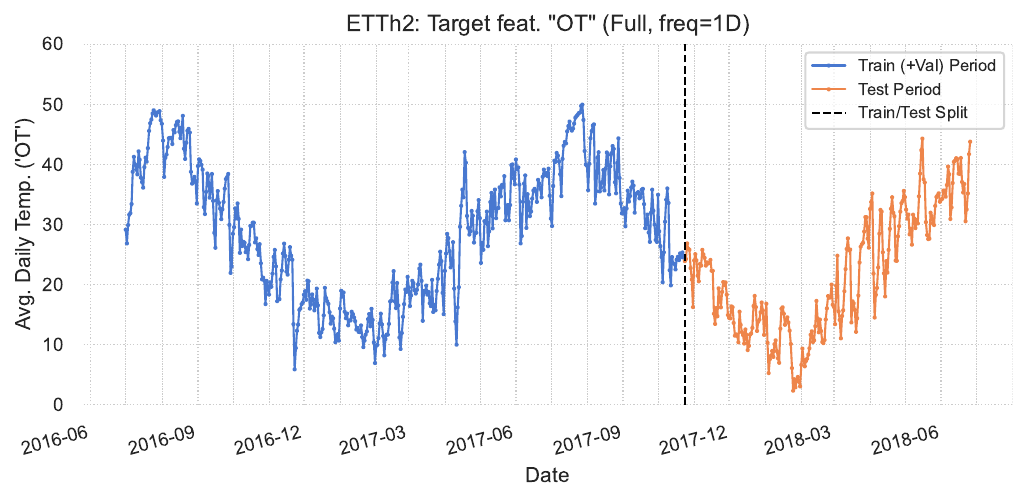}
        \end{subfigure} &
        \begin{subfigure}[b]{0.19\textwidth}
            \centering
            \includegraphics[width=\linewidth]{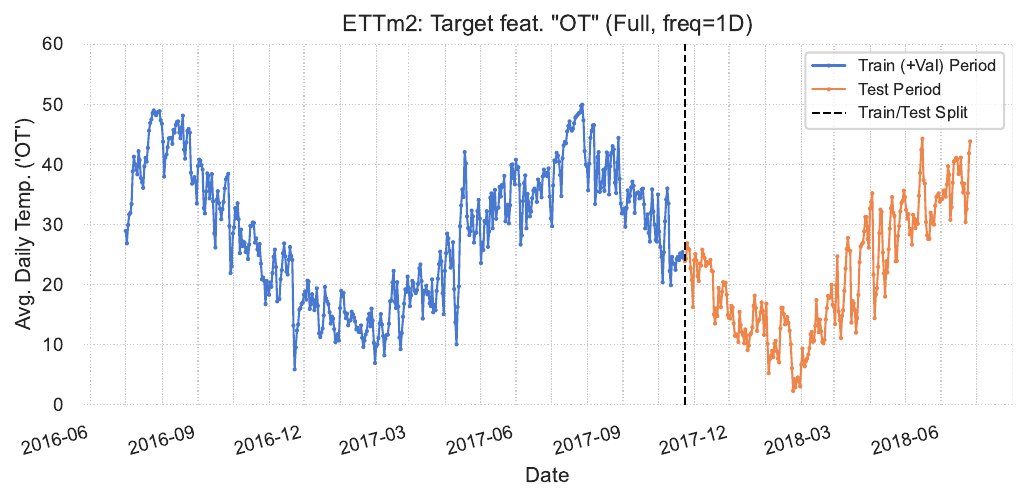}
        \end{subfigure} &
        \begin{subfigure}[b]{0.19\textwidth}
            \centering
            \includegraphics[width=\linewidth]{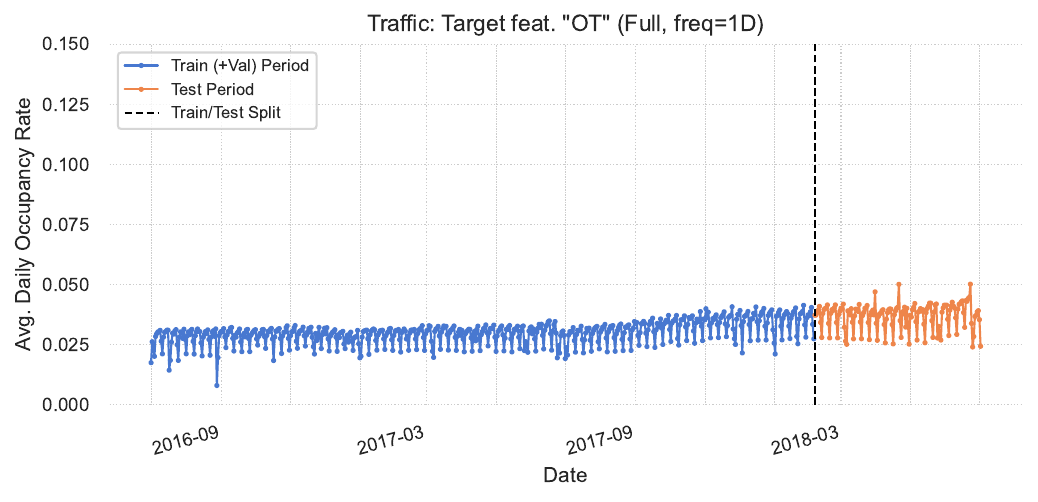}
        \end{subfigure} &
        \begin{subfigure}[b]{0.19\textwidth}
            \centering
            \includegraphics[width=\linewidth]{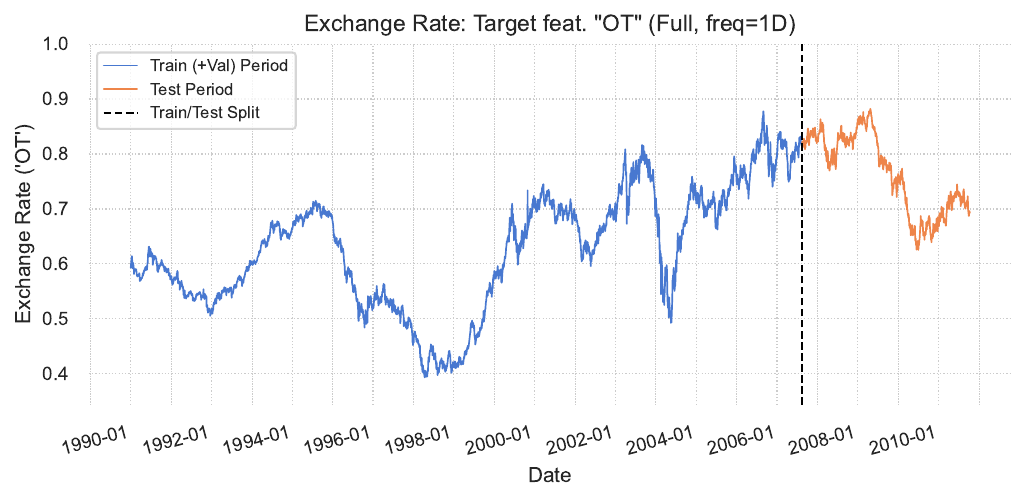}
        \end{subfigure} &
        \begin{subfigure}[b]{0.19\textwidth}
            \centering
            \includegraphics[width=\linewidth]{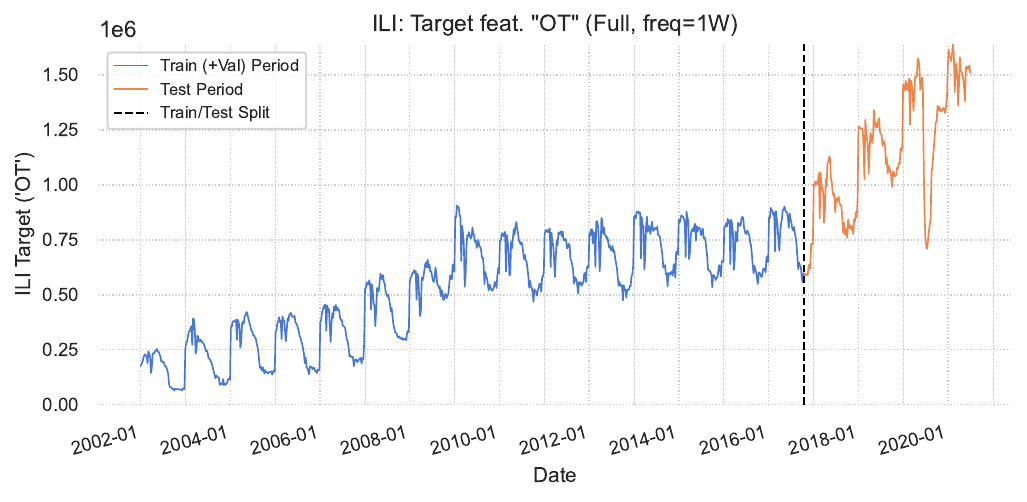}
        \end{subfigure} \\

        \begin{subfigure}[b]{0.19\textwidth}
            \centering
            \includegraphics[width=\linewidth]{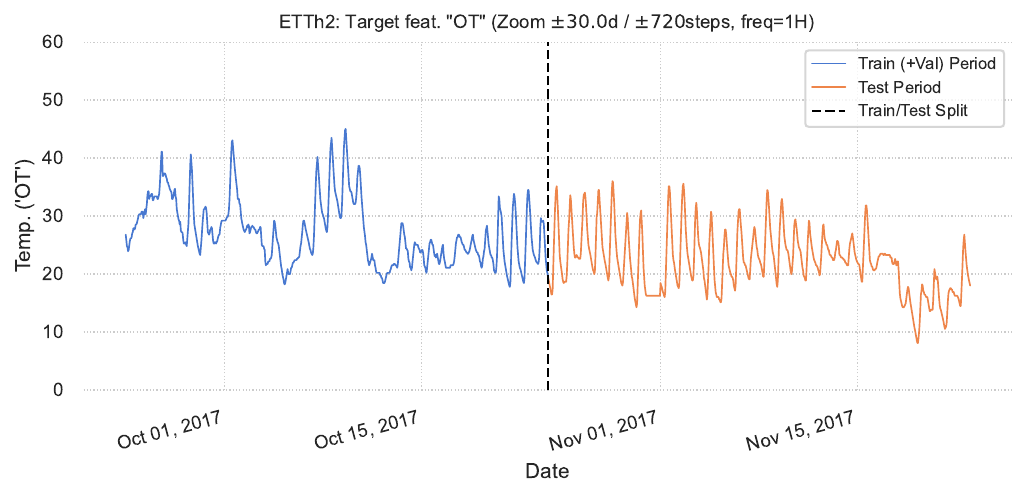}
            \caption*{\tiny(a) \textsc{ETT}h2 `OT'}
        \end{subfigure} &
        \begin{subfigure}[b]{0.19\textwidth}
            \centering
            \includegraphics[width=\linewidth]{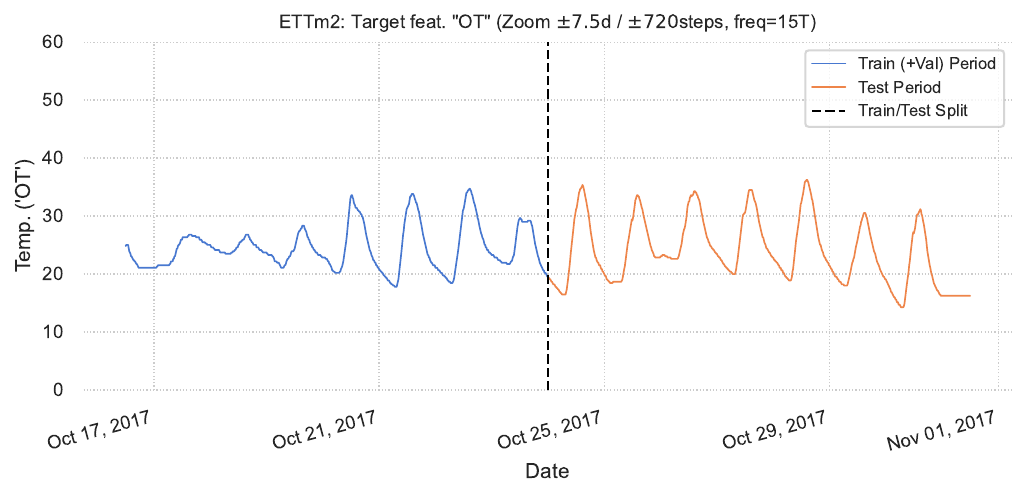}
            \caption*{\tiny(b) \textsc{ETT}m2 `OT'}
        \end{subfigure} &
        \begin{subfigure}[b]{0.19\textwidth}
            \centering
            \includegraphics[width=\linewidth]{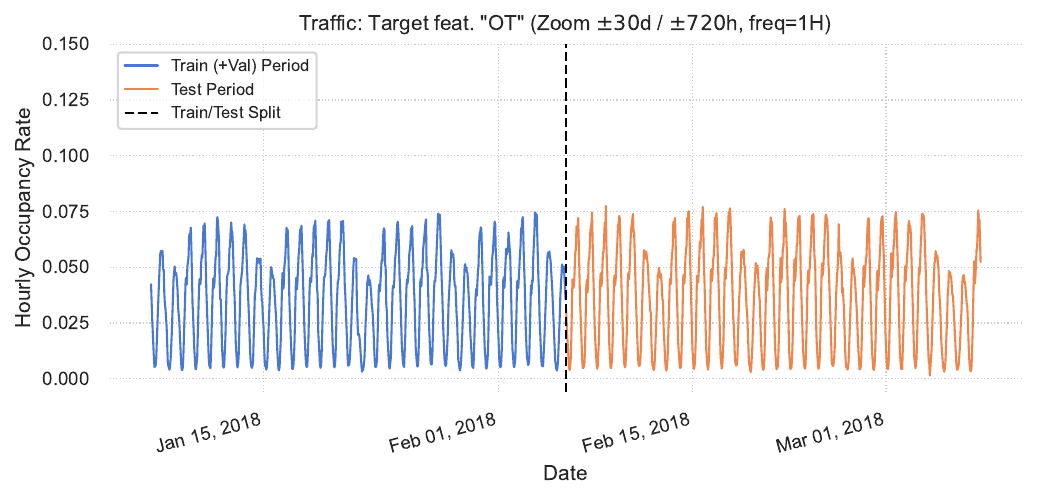}
            \caption*{\tiny(c) \textsc{Traffic} `OT'}
        \end{subfigure} &
        \begin{subfigure}[b]{0.19\textwidth}
            \centering
            \includegraphics[width=\linewidth]{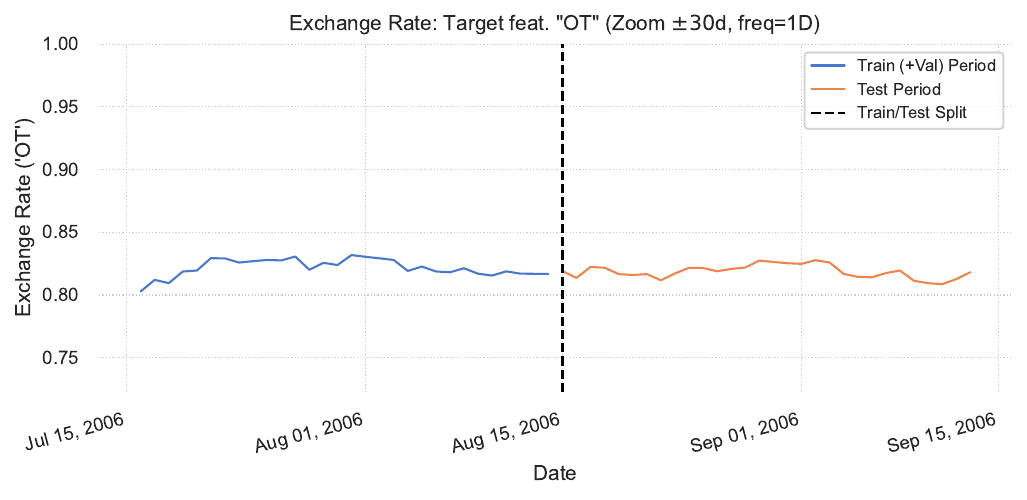}
            \caption*{\tiny(d) \textsc{Exchange} `OT'}
        \end{subfigure} &
        \begin{subfigure}[b]{0.19\textwidth}
            \centering
            \includegraphics[width=\linewidth]{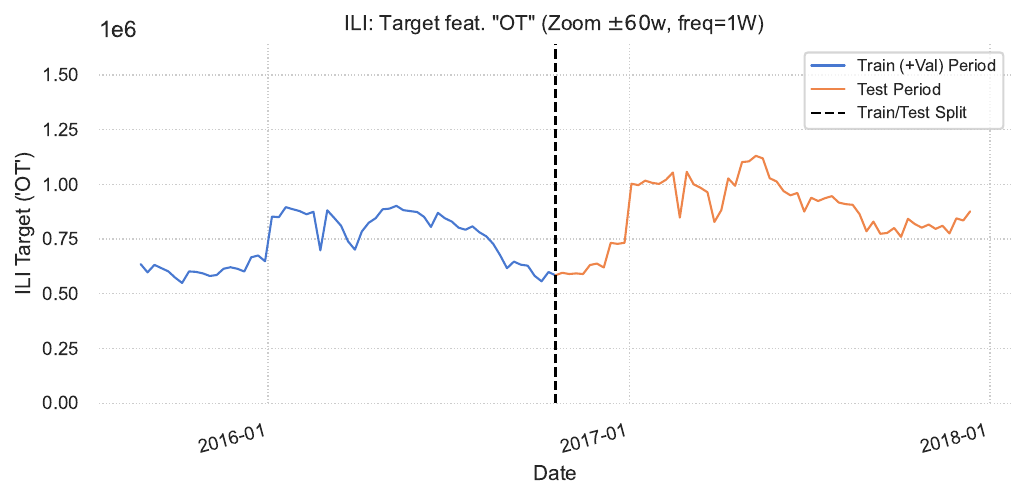}
            \caption*{\tiny(e) \textsc{ILI} `OT'}
        \end{subfigure}
    \end{tabular}
    \caption{Comparative visualization of the \textbf{target variable} (consistently named `OT' in these preprocessed versions) dynamics across five key LTSF benchmark datasets. \textbf{Top row:} Full time series plotted with daily frequency (weekly for \textsc{ILI}) to highlight long-term trends and seasonalities. The \emph{vertical dashed line} demarcates the \colorwb{\textbf{train data (left)}} and \textcolor{orange}{\textbf{test data (right)}}.
    \textbf{Bottom row:} Zoomed-in views ($\pm$ largest typical forecast horizon: $\pm 30$ days for hourly \textsc{ETT}h2; $\pm 7.5$ days for 15-min \textsc{ETT}m2; $\pm 30$ days for hourly \textsc{Traffic}; $\pm 30$ days for daily \textsc{Exchange}; and $\pm 60$ weeks for weekly \textsc{ILI}) around the train/test split, plotted at their original data frequency. These plots reveal the nature and persistence of inherent \colorwblk{periodic patterns} across the \colorwb{train}/\textcolor{orange}{test} boundary.}
    \label{fig:ltsf_dataset_periodicity}
\end{figure*}

Following the rapid progression in the language model (LM) paradigm—from pretrained LMs (PLMs)~\cite{zhao2023survey} to increasingly larger generative LLMs and instruction-aligned chatbots~\cite{tsf_survey_fm_zhou2023comprehensive}—the breadth and taxonomic complexity of AI/ML-based TSF models have consequently ballooned. In unison, the purported novelty of these increasingly complex models, and by extension their acceptance at prestigious AI/ML conferences, has largely rested on achieving meagre decimal-point Mean Squared Error (MSE) improvements over these LTSF datasets. Yet, paradoxically, these same datasets often yield competitive, or even superior, results when modeled with vanilla statistical models like AR, ARIMA~(\citeyear{stats_ar_arma_whittle1951hypothesis, arima_shumway2017time}) or using well-traversed techniques from the decades-old general TSF research toolbox~(\citeyear{holt-winters_kalekar2004time, tsf_auto-arima_hyndman2008automatic}). This disparity raises serious questions about the incremental value and justification for the burgeoning complexity of many deep learning approaches in these contexts.

We are not the first to highlight this discrepancy. A 2022 critique by Zeng et al.~\cite{tsf_critique_zeng2023transformers} underscored this issue by introducing a suite of remarkably simple \colorwblkb{one-layer linear models}, termed LTSF-Linear, as baseline comparators. Surprisingly, across all nine standard LTSF benchmark datasets, LTSF-Linear consistently outperformed the then state-of-the-art (SOTA) transformer-based models—often by a substantial margin. Subsequent works on \textit{ultra-light models}—notably SparseTSF~\cite{tsf_sparsetsf-lin24n} with fewer than \emph{1k} parameters—have further demonstrated competitive or superior SOTA performance on these benchmarks.

Delving into the mechanics of models like SparseTSF offers insights into their surprising effectiveness. At a high level, techniques such as SparseTSF effectively decompose a time series $X(t)$ into its periodic component $P(t)$ (where $P(t) = P(t + w)$ for some period $w$) and a trend component $T(t)$, such that $X(t) = P(t) + T(t)$. By isolating or accounting for the highly regular $P(t)$, the model is left to primarily learn a mapping function $f:X(t) \rightarrow T(t)$, a task often achievable with a simple MLP layer. For such straightforward techniques to be effective, however, the crucial prerequisite is the existence of a strong, persistent periodic component in the data. As it transpires, the ``diverse'' LTSF datasets largely fulfill this condition. Figure~\ref{fig:ltsf_dataset_periodicity} visually confirms the strong, consistent periodicities (e.g., daily, weekly, annual) in the target variables of these datasets, patterns that persist across their train and test splits. Intuitively, datasets derived from electricity usage, traffic patterns, or weather phenomena are inherently seasonal.

While these datasets certainly represent diverse source domains, their dominant characteristic of strong periodicity means they occupy a relatively narrow band within the broader spectrum of time series non-stationarities (which includes complexities like structural breaks, concept drift, and time-varying volatility, cf. Figure~\ref{fig:ts_nonstationarity}, Table~\ref{tab:ts_nonstationarity} in Appendix~\S\ref{app:ssec:non-stationarities}). The current benchmarking practice, therefore, risks \textbf{overstating model generalization capabilities} if performance is primarily assessed on this limited type of series.

Despite a slew of critical works alluding to these disparities, even in 2025, we continue to observe transformer-variants, billion-parameter foundation models, and LLMs being applied to these same benchmark datasets—almost invariably without rigorous comparison to appropriate na\"{i}ve or classical statistical baselines. If the community persists in benchmarking predominantly on data with trivial-to-model periodicities while omitting robust classical baselines, we risk an echo chamber effect: chasing marginal improvements rather than addressing the pressing challenges of non-stationarity, distribution drift, and real-world volatility that underpin critical applications in finance, energy, and healthcare.

This paper highlights these fundamental shortcomings in current TSF benchmarking practices and offers concrete recommendations—grounded in decades of well-established forecasting research—for constructing more demanding, task-relevant datasets and evaluation protocols that mandate the inclusion of cheap, classical baselines.

\paragraph{Organization:} The paper is organized as follows: \S\ref{sec:background} provides abridged preliminaries; \S\ref{sec:llm_tsf_challenges} outlines challenges for deep learning models in TSF; \S\ref{sec:critique} and \S\ref{sec:critique-benchmarking} discuss pitfalls in current evaluation schemes; and \S\ref{sec:conclusion} concludes the paper.

\vspace{1em}
\section{Preliminaries and Background}\label{sec:background}

We begin by formally defining the time series forecasting (TSF) problem. At an intuitive level, time series observations are realizations (or sample paths) from an underlying, often unobserved, stochastic process, typically recorded at discrete time intervals.

\paragraph{Point Forecasting Problem.}
The fundamental task of \textit{univariate point forecasting in discrete time} is to predict a vector of future values, $\hat{y} \in \mathbb{R}^H = [\hat{y}_{T+1}, \hat{y}_{T+2}, \dots, \hat{y}_{T+H}]$, over a specified forecast horizon $H$ (also known as lead time). This prediction is conditioned on an observed history of $T$ past values of the series, denoted as $\{y_1, \dots, y_T\} \in \mathbb{R}^T$.
The \textbf{multivariate forecasting} variant extends this by conditioning the prediction $\hat{y}$ not only on the past of the target series but also on the history of $d-1$ other related time series, or exogenous variables. Thus, the input can be represented as $X \in \mathbb{R}^{d \times T'}$ (where $T'$ is the lookback window length), and the output remains $\hat{y} \in \mathbb{R}^{H}$ for a single target series, or $\hat{Y} \in \mathbb{R}^{d' \times H}$ if forecasting multiple $d'$ series simultaneously.

\paragraph{Model Categories.}
AI/ML-based TSF models can be broadly categorized, though taxonomies continue to evolve:
\textbf{(i) Specialist Transformer-based Models:} Architectures like Informer~\cite{tsf_zhou2021informer}, Autoformer~\cite{tsf_wu2021autoformer}, FEDformer~\cite{tsf_zhou2022fedformer}, Pyraformer~\cite{tsf_liu2021pyraformer}, and PatchTST~\cite{tsf_patchtst_nie2022time} are purpose-built for TSF. They are typically trained from scratch on specific TSF tasks and are explicitly designed to model temporal dependencies.
\textbf{(ii) Specialist Multimodal TSF LLMs/FMs:} These are Foundation Models (FMs)~\cite{bommasani2021opportunities}, often based on LLM architectures, that incorporate diverse time-series datasets, potentially alongside other modalities (e.g., text, images). They are either trained fully for TSF or fine-tuned from general-purpose pretrained models, aiming to leverage cross-modal signals for improved domain-specific forecasting~\cite{moirai-woo24a, tsf_aaai25_liu2025calf}.
\textbf{(iii) Generalist LLMs:} Large Language Models such as ChatGPT~\cite{li2023chatgpt} and Gemini~\cite{anil2023gemini}, pretrained on vast internet-scale corpora, are sometimes applied to TSF, typically through prompting or lightweight adaptation, leveraging their general pattern recognition and reasoning capabilities.

\paragraph{Evaluation Metrics.}
While numerous metrics exist for TSF evaluation (see Appendix~\ref{app:metrics}), our discussion primarily focuses on two prevalent in AI/ML research: Mean Squared Error (MSE) and Mean Absolute Error (MAE). These are defined as:
\begin{equation}\label{eq:metrics}
    \centering
    \text{MSE} = \frac{1}{H} \sum_{i=1}^{H} ( y_{T+i} - \hat{y}_{T+i} )^2 \quad \text{(a)} \qquad \qquad
    \text{MAE} = \frac{1}{H} \sum_{i=1}^{H} | y_{T+i} - \hat{y}_{T+i} | \quad \text{(b)}
\end{equation}
where $H$ is the forecast horizon, $y_{T+i}$ is the true value at time $T+i$, and $\hat{y}_{T+i}$ is the model's prediction. Both MSE and MAE aggregate point-wise errors over the forecast window. Their sensitivity to data scale makes them straightforward for optimizing individual series but less suitable for direct cross-series performance comparisons without normalization.

\begin{wrapfigure}[14]{r}{0.50\textwidth}
    \centering
    \vspace{-1.2em} 
    \includegraphics[width=1\linewidth]{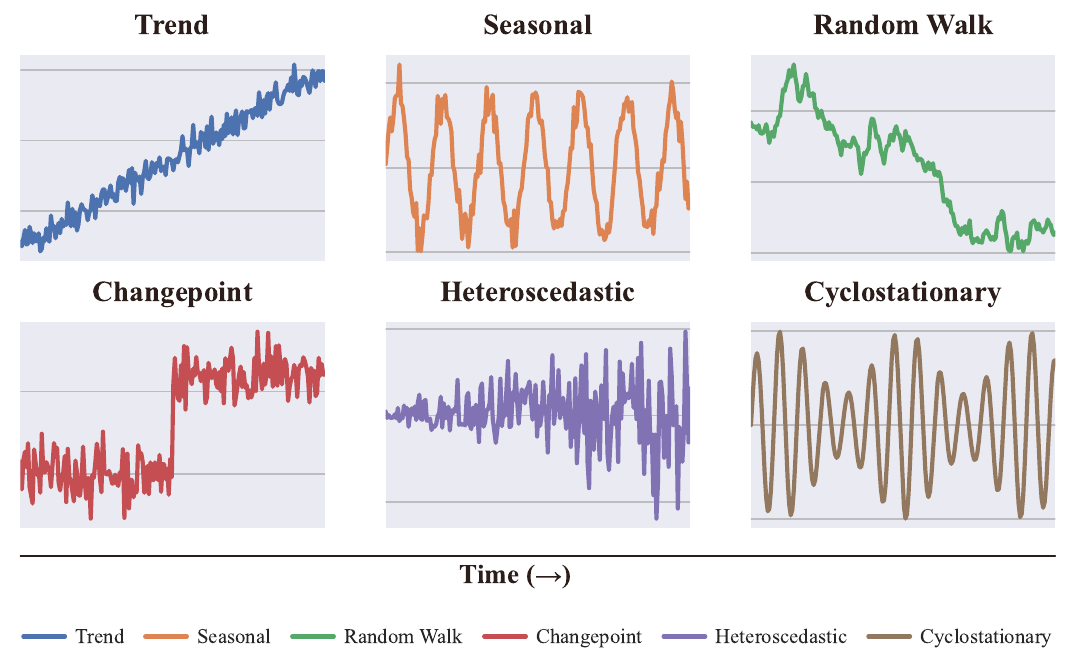}
    \caption{Schematic illustration of canonical non-stationarities commonly observed in time series data.}
    \label{fig:ts_nonstationarity}
    \vspace{-1.2em}
\end{wrapfigure}

\paragraph{Spectrum of Time Series Characteristics.}
Real-world time series data can exhibit a multitude of characteristics, including various forms of non-stationarity~\cite{box2015time}, as schematically illustrated in Figure~\ref{fig:ts_nonstationarity}.

A more detailed taxonomy of these characteristics is provided in Appendix~\ref{app:ssec:non-stationarities} (Table~\ref{tab:ts_nonstationarity}). The crucial implication is that \colorwblk{time series data possess diverse statistical properties}. Consequently, appropriate baseline methodologies and evaluation metrics are inherently nuanced and must be conditioned on the specific domain, task, and observed data characteristics.
This paper critically examines current benchmarking practices that often overlook this diversity, particularly in the context of LTSF tasks, by predominantly utilizing datasets with a narrow range of these characteristics (e.g., strong, persistent periodicity).

\section{General Challenges of LLMs in Time-Series Forecasting}
\label{sec:llm_tsf_challenges}

LLMs have recently been explored for their potential application in time-series forecasting (TSF) across multiple domains; including finance, healthcare, and economics. However, despite their success in language-related tasks, LLMs present significant challenges when applied to TSF. In this section, we critically assess these challenges, drawing from recent studies that provide an evaluation of LLM performance in TSF tasks.

Reasoning about \textit{temporal data} effectively remains a fundamental challenge with LLMs in TSF. In their study, Merrill et al.~(\citeyear{tsf_merrill2024language}) argue that LLMs perform poorly in zero-shot reasoning for time-series tasks, such as etiological reasoning, question answering, and context-aided forecasting. Etiological reasoning, which involves hypothesizing potential causes for a given time series, showed that even strong LLMs like GPT-4~\cite{openai2023gpt4} performed marginally better than \colorwblkb{random guessing} and lagged significantly behind human performance by as much as 30 percentage points. Similarly, question-answering tasks involving time-series data also demonstrated the limitation of LLMs to adequately interpret and relate temporal information, resulting in scores that were at or \colorwblkb{near random} in many cases. This challenge and the empirical limitations highlight an important question: might the transformer architecture~\cite{vaswani2017attention} --- the default underlying structural building block underpinning most LLMs --- itself be inherently ill-suited for modeling temporal dynamics in time-series tasks?

Zeng et al.~\cite{tsf_critique_zeng2023transformers} explored this very question of the effectiveness of transformer-based models for TSF. Self-attention without an order-aware mechanism is permutation-equivariant and therefore does not by itself represent temporal order. To illustrate the practical concern, they proposed a class of embarrassingly simple models (LTSF-Linear) --- including a one-layer linear model (DLinear) --- that empirically outperformed more complex transformer-based models on the LTSF benchmark datasets. Their results suggest that the inductive biases and complexity of many Transformer designs may be poorly matched to these particular TSF tasks.

This argument should not be read as showing that Transformers necessarily discard all positional information. Haviv et al.~\cite{haviv2022transformer} demonstrate that causal language models without explicit positional encodings can learn a useful notion of absolute position, plausibly because the causal mask reveals how many predecessors each token can attend to. Their result concerns causal language modeling rather than TSF, but it shows that explicit positional encodings are not the only source of order information. The pertinent forecasting question is therefore empirical: whether a model's order-aware mechanisms and inductive biases capture the temporal dynamics of the target task well enough to justify their added complexity.

In sum, this critique leads to a deeper, arguably more fundamental concern: whether the added complexity and computational cost of LLMs are ever truly justified by a \colorwblkb{commensurate gain} in TSF performance.

Echoing this sentiment, Tan et al.~(\citeyear{tsf_critique_tan2024language}) argue that LLMs offer marginal, if any, improvements in forecasting accuracy when \underline{compared to far simpler, cheaper alternatives}. Their ablation studies revealed that replacing LLM components with basic attention layers or even removing them altogether often improved performance or left it unaffected, while drastically reducing computational requirements. The findings indicate that despite the advanced architectures of LLMs, they fail to represent the sequential dependencies in time-series data effectively, particularly for forecasting applications. Moreover, pretrained LLMs did not provide any significant advantage in few-shot settings, which underscores the limitations of their sequence modeling capabilities in non-language contexts.

\noindent These studies collectively underscore the limitations of current LLMs and transformer-based models when applied to time-series forecasting. The \textbf{key weaknesses} can be \textbf{summarized} as follows:

\vspace{0.5em}
\begin{tcolorbox}[colback=red!5!white, colframe=black, width=\textwidth, boxsep=4pt, arc=4pt]

    \begin{itemize}[leftmargin=*, labelindent=0pt, itemsep=6pt, topsep=2pt, parsep=0pt]
        \item[\textcolor{red!80!black}{\ding{115}}] \textbf{Limited Temporal Reasoning and Causal Understanding:}
        LLMs perform poorly in zero-shot settings involving etiological reasoning and temporal question answering, often scoring only marginally above random~\cite{tsf_merrill2024language}.

        \item[\textcolor{red!80!black}{\ding{115}}] \textbf{Temporal Ordering Requires Appropriate Inductive Biases:}
        Self-attention without order-aware structure is permutation-equivariant. Explicit positional encodings or causal masks can supply position information, but benchmark evidence still shows that standard Transformer designs may be poorly matched to some TSF tasks~\cite{tsf_critique_zeng2023transformers, haviv2022transformer}.

        \item[\textcolor{red!80!black}{\ding{115}}] \textbf{Underutilization of Contextual Information:}
        Even when supplied with external context (e.g., anticipated events, trend descriptions), LLMs show limited capacity to adjust forecasts meaningfully, indicating shallow integration of non-observed signals~\cite{tsf_merrill2024language}.

        \item[\textcolor{red!80!black}{\ding{115}}] \textbf{Overfitting and Error Accumulation in Long-Horizon Forecasting:}
        Transformer-based models often overfit short-term patterns, leading to compounding errors across long horizons—particularly problematic in dynamic environments with distributional drift or structural change~\cite{tsf_critique_zeng2023transformers}.

        \item[\textcolor{red!80!black}{\ding{115}}] \textbf{High Computational Cost Without Commensurate Gains:}
        Ablation studies reveal that the architectural complexity of LLMs often fails to translate into improved forecasting performance. In some cases, removing or simplifying LLM components even improves accuracy while drastically reducing computational overhead~\cite{tsf_critique_zeng2023transformers, tsf_critique_tan2024language}.
    \end{itemize}
\end{tcolorbox}
\vspace{0.5em}

\noindent Taken together, these findings cast doubt on the suitability of current LLMs and large-scale transformers for TSF tasks. Future research must prioritize models that can natively encode temporal dependencies, integrate dynamic external context, and generalize under regime shifts—while maintaining computational efficiency and interpretability.

\section{A Thought Experiment Aided Critique of TSF LLMs}\label{sec:critique}

\subsection{The Seeking SOTA Experiment}\label{ssec:seeking_sota_main}
To intuitively illustrate the pitfalls of current evaluation paradigms, particularly how aggregated metrics can favor generalist consistency over specialist excellence, we employ a thought experiment. In this constructed scenario, one of the authors is posited as a candidate for the ``best overall human athlete'' by evaluating performance across a diverse set of four sporting disciplines: running, cricket, soccer, and chess. These disciplines, much like the varied datasets in TSF benchmarks, possess different characteristics, units of measurement, and require distinct skill sets.

The core of the experiment, detailed in Appendix~\S\ref{app:ssec:seeking-sota} along with the constructed performance data (Table~\ref{tab:thought_experiment}), demonstrates a crucial point: despite the author being demonstrably inferior to world-class specialists in their respective primary sports (e.g., Usain Bolt in sprinting, Magnus Carlsen in chess), the author can achieve the highest overall "weighted average" score. This outcome is engineered by assuming the author maintains a consistently high (e.g., >90th percentile) performance across all four disparate sports—a strong generalist capability—and by leveraging an evaluation scheme that, through specific weighting or metric scaling, disproportionately rewards such broad competence. For instance, as shown in Table~\ref{tab:thought_experiment} (Appendix~\S\ref{app:ssec:seeking-sota}), the author might achieve a top percentile in a less conventionally weighted metric (like soccer assists per game), while specialists, despite their world-record performances in one area, show significantly lower capabilities in others.

This \colorwblkb{Seeking SOTA Athlete} analogy serves to highlight how global TSF models might achieve benchmark SOTA not by universally outperforming specialized models on individual tasks, but by performing consistently well across a collection of diverse datasets, especially if the evaluation scheme averages performance or emphasizes generalizability. The contrived victory of the "generalist author" underscores our argument that aggregated SOTA can be misleading, potentially obscuring true utility for specific tasks and creating an illusion of superior overall capability based on the benchmark's particular construction and aggregation method.

\subsection{Are TSF LLMs Secretly Time Series \colorwblk{Stein Estimators}?}
\label{ssec:stein}

The ``Seeking SOTA Experiment'' (\S\ref{ssec:seeking_sota_main}, Appendix~\S\ref{app:ssec:seeking-sota}) offers an intuitive parallel to \colorwblk{Stein's Paradox}~\cite{efron1977stein, stein_feldman2014revisiting} and its implications for evaluating modern Time Series Forecasting (TSF) Foundation Models (FMs). Stein's Paradox reveals that when simultaneously estimating multiple (three or more) parameters, a combined estimator, like the James-Stein estimator~\cite{stein_james1961estimation}, can achieve a lower total squared error risk than estimating each parameter independently. This improvement occurs by ``shrinking'' individual estimates towards a common mean, effectively borrowing strength across observations, even if the parameters seem unrelated~\cite{efron1977stein, stein_samworth2012stein}.

This statistical principle resonates strongly with the behavior of global TSF FMs. These models, often based on Transformer~\cite{tsf_gpt4ts_neurips23_zhou2023one} or Mixer~\cite{tsf_ttms_ekambaram2024neurips} architectures, are trained on vast collections of diverse time series (e.g., the LTSF datasets, Table~\ref{tab:ltsf_ett_mse}). They learn shared representations by minimizing an average error metric across all constituent series, analogous to how Stein estimators operate. By learning from a multitude of series, these FMs implicitly regularize or ``shrink'' the effective model for any individual series towards a generalized structure learned from the entire corpus. This aligns with concepts like multi-task averaging (MTA) as an empirical Bayes strategy~\cite{stein_feldman2014revisiting} and the ``globality'' principle in TSF, where leveraging information across series can improve average accuracy, especially for noisy or short series~\cite{montero2021principles}.

The paradox illustrated by our thought experiment --- where the generalist author achieves the highest aggregate score despite specialist superiority in individual tasks --- mirrors how a global TSF model might attain benchmark SOTA. Such a model could outperform specialized local models in overall rankings not by universal task-specific dominance, but by consistent ``good enough'' performance across many diverse series, particularly if evaluation relies on averaged metrics. This raises a \colorwblkb{critical question}: does such aggregated SOTA reflect true utility for specific \colorwblk{practical forecasting applications}, or merely proficiency at the benchmark's particular aggregation scheme, potentially sacrificing peak performance on individual series?

For instance, on the LTSF benchmark with disparate datasets like ETT~\cite{tsf_dataset_electricity} and \textsc{ExchangeRate}~\cite{tsf_dataset_exchange_lai2018modeling}, a global model might benefit from Stein-like shrinkage, learning transferable patterns. However, this ``averaged'' excellence, much like the author's claimed (Pyrrhic at best, egregious at worst) victory, could obscure the potential superiority of a specialized, ``local'' model for a single critical task. The James-Stein estimator can increase risk for individual parameters far from the shrinkage target. Similarly, blindly optimizing global TSF models for average performance across heterogeneous datasets might yield ``masters of the average, but not of any single instance,'' potentially compromising peak performance where it is most needed. This underscores the necessity for nuanced evaluation protocols in TSF that balance generalizability with the demand for specialized accuracy.

\section{The State of Benchmarking in AI/ML TSF Models}
\label{sec:critique-benchmarking}
While the oft-quoted Goodhart's Law serves as a cautionary reminder in empirical research, the current benchmarking landscape in the AI/ML TSF models in the specialist transformer-variants category --- or \textit{local models} as per time series research terminology~\cite{montero2021principles} ---  remains so ill-defined that even the risk of the law manifesting seems premature! We have a more foundational problem at hand with the absence of standardized, scientifically rigorous, principled evaluation benchmark as illustrated by the scores of top AI/ML conference works\footnote{Our critique is directed at evaluation protocols, benchmark composition, and reporting conventions rather than at the intent or competence of authors of prior work. Named papers (if applicable) are discussed only as representative examples of broader methodological patterns.}
 benchmarking on the so-called standard LTSF datasets listed in Table~\ref{tab:ltsf_ett_mse}.

\paragraph{\ding{42}~The Danger of Cascading Errors}
Ironically, our overarching call to action --- advocating for task-specific, standardized, third-party (i.e. neutral), taxonomized TSF benchmarks --- stems from noted challenges that are \textbf{not} by any means \textbf{novel}. 
Numerous contemporaneous studies have already critiqued the illusion of progress~\cite{tsf_critique_zeng2023transformers, tsf_critique_tan2024language},
and documented recurring evaluation pitfalls~\cite{tsf_critique_hewamalage2023forecast}. 
These concerns remain influential because benchmark choices and reporting conventions introduced in highly cited papers often persist into subsequent work. For example, Informer~(AAAI~\citeyear{tsf_zhou2021informer}) helped establish the LTSF benchmark that later became widely used.
More recent examples include TimeLLM~(ICLR~\citeyear{tsf_timellm_iclr24_jin2023time}) and CALF~(AAAI~\citeyear{tsf_aaai25_liu2025calf}).
TimeLLM has been discussed in subsequent studies (with over 500 citations) for discrepancies between reported results and later independent reproductions~\cite{tsf_ss-ip-llm_pan2024textbf, tsf_ttms_ekambaram2024ttms, tsf_fm_moment_goswami2024momentfamilyopentimeseries, tsf_critique_xu2025specialized}. Nonetheless, it continues to be used as a baseline in later work, including CALF, which reuses its official codebase and experimental setup with the stated aim of maintaining evaluation consistency.
More broadly, we observe reporting choices that can reduce interpretability across papers. For instance, TimeLLM places full horizon-specific LTSF results in the appendix and emphasizes aggregated average MSE and MAE across horizons, which can obscure performance variability across forecast lengths (~\S\ref{sssec:improper-metric},~\S\ref{app:limitations-mse-mae}). This issue is especially relevant for LLM-style TSF global models~\cite{montero2021principles}, where aggregate gains may mask weaker performance at a more granular level. CALF follows a similar reporting pattern and does not include the full results table in the main manuscript.

\subsection{Challenges, Weaknesses in Evaluation}

\begin{table}[!htbp]
\centering
\caption{Long‐term forecasting \textbf{MSE} ($\downarrow$) results on four ETT datasets for horizons $\{96,192,336,720\}$. Models are ordered by (descending) publication year. The specialist transformer variant model results are reproduced as is from \textit{Table~16} of~\cite{tsf_wang2025timemixer++} without reimplementation by the authors. \colorwr{Discrepancies} with \colorwg{original publication values*} are \colorwr{highlighted}. The TSF foundation model (FM) results are obtained from \textit{Table~14} of \citet{tsf_critique_xu2025specialized}. \textbf{Best results} are in \textbf{bold}, closest to best are \underline{underlined} for emphasis. $\spadesuit$ denote reimplementation by the authors.
}
\label{tab:ltsf_ett_mse}
\vspace{1ex}
\renewcommand{\arraystretch}{1.1}
\rowcolors{2}{gray!5}{white}
\small
\resizebox{\linewidth}{!}{
  \begin{tabular}{@{}l
                  *{4}{rrrr}
                  *{4}{rrrr}
                  @{}}
    \toprule
    \multirow{2}{*}{\textbf{Model (Year)}}
      & \multicolumn{4}{c}{\textbf{ETTh1}}
      & \multicolumn{4}{c}{\textbf{ETTh2}}
      & \multicolumn{4}{c}{\textbf{ETTm1}}
      & \multicolumn{4}{c}{\textbf{ETTm2}} \\
    \cmidrule(lr){2-5} \cmidrule(lr){6-9} \cmidrule(lr){10-13} \cmidrule(l){14-17}
      & 96  & 192 & 336 & 720
      & 96  & 192 & 336 & 720
      & 96  & 192 & 336 & 720
      & 96  & 192 & 336 & 720 \\
    \midrule

    Informer~(\citeyear{tsf_zhou2021informer})
                                  & 0.865 & 1.008 & 1.107 & 1.181
                                  & 3.755 & 5.602 & 4.721 & 3.647
                                  & -     & -     & -     & -
                                  & -     & -     & -     & - \\

    Autoformer~(\citeyear{tsf_wu2021autoformer})
                                  & 0.449 & 0.500 & 0.521 & 0.514
                                  & 0.346 & 0.456 & 0.482 & 0.515
                                  & 0.505 & 0.553 & 0.621 & 0.671
                                  & 0.255 & 0.281 & 0.339 & 0.433 \\
    Stationary~(\citeyear{tsf_non-stationary.transformers_liu2022non_neurips2022})
                                  & 0.513 & 0.534 & 0.588 & 0.643
                                  & 0.476 & 0.512 & 0.552 & 0.562
                                  & 0.386 & 0.459 & 0.495 & 0.585
                                  & 0.192 & 0.280 & 0.334 & 0.417 \\
    FEDformer~(\citeyear{tsf_zhou2022fedformer})
                                  & 0.395 & 0.469 & 0.530 & 0.598
                                  & 0.358 & 0.429 & 0.496 & 0.463
                                  & 0.379 & 0.426 & 0.445 & 0.543
                                  & 0.203 & 0.269 & 0.325 & 0.421 \\
    SCINet~(\citeyear{tsf_liu2022scinet_neurips2022})
                                  & 0.654 & 0.719 & 0.778 & 0.836
                                  & 0.707 & 0.860 & 1.000 & 1.249
                                  & 0.418 & 0.439 & 0.490 & 0.595
                                  & 0.286 & 0.399 & 0.637 & 0.960 \\

    TimesNet~(\citeyear{tsf_timesnet})
                                & 0.384 & 0.436 & 0.491 & 0.521
                                & 0.340 & 0.402 & 0.452 & 0.462
                                & 0.338 & 0.374 & 0.410 & 0.478
                                & 0.187 & 0.249 & 0.321 & 0.408 \\
    TiDE~(\citeyear{tsf_tide_das2023long})
                                & 0.479 & 0.525 & 0.565 & 0.594
                                & 0.400 & 0.528 & 0.643 & 0.874
                                & 0.364 & 0.398 & 0.428 & 0.487
                                & 0.207 & 0.290 & 0.377 & 0.558 \\
    Crossformer~(\citeyear{tsf_crossformer_zhang2023_iclr2023})
                                & 0.423 & 0.471 & 0.570 & 0.653
                                & 0.745 & 0.877 & 1.043 & 1.104
                                & 0.404 & 0.402 & 0.484 & 0.666
                                & 0.287 & 0.414 & 0.597 & 1.730 \\
    PatchTST~(\citeyear{tsf_patchtst_nie2022time})
                                & 0.460 & 0.477 & 0.546 & 0.544
                                & 0.308 & 0.393 & 0.427 & 0.436
                                & 0.352 & 0.374 & 0.421 & 0.462
                                & 0.183 & 0.255 & 0.309 & 0.412 \\

    iTransformer~(\citeyear{tsf_liu2023itransformer_iclr2024})
                                & 0.386 & 0.441 & 0.487 & 0.503
                                & 0.297 & 0.380 & 0.428 & 0.427
                                & 0.334 & 0.390 & 0.426 & 0.491
                                & 0.180 & 0.250 & 0.311 & 0.412 \\

    \rowcolor{warmgreen!20}
    SparseTSF*~(\citeyear{tsf_sparsetsf-lin24n})
                                  & 0.359 & 0.397 & 0.404 & 0.417
                                  & 0.267 & 0.314 & 0.312 & 0.307
                                  & -     & -     & -     & -
                                  & -     & -     & -     & - \\
    \rowcolor{warmblue!20}
    SparseTSF~$\spadesuit$~(\textit{reimpl. by us})
                            & 0.363 & 0.400 & 0.435 & 0.424
                            & 0.295  & 0.340  & 0.360  & 0.383
                            & -     & -     & -     & -
                            & -     & -     & -     & - \\

    TimeMixer~(\citeyear{tsf_wang2024timemixer})
                                & 0.375 & 0.429 & 0.484 & 0.498
                                & 0.289 & 0.372 & 0.386 & 0.412
                                & 0.320 & 0.361 & 0.390 & 0.454
                                & 0.175 & 0.237 & 0.298 & 0.391 \\
    \rowcolor{warmyellow}
    TimeMixer\texttt{++}~(\citeyear{tsf_wang2025timemixer++})
                    & 0.361 & 0.416 & 0.430 & 0.467
                    & 0.276 & 0.342 & 0.346 & 0.392
                    & 0.310 & 0.348 & 0.376 & 0.440
                    & 0.170 & 0.229 & 0.303 & 0.373 \\

    \midrule

    GPT4TS (OFA)~(\citeyear{tsf_gpt4ts_neurips23_zhou2023one})
                         & 0.376 & 0.416 & 0.442 & 0.477
                         & 0.285 & 0.354 & 0.373 & 0.406
                         & 0.292 & 0.332 & 0.366 & 0.417
                         & 0.173 & 0.229 & 0.286 & 0.378 \\
    TEST (Few shot)~(\citeyear{tsf_test_sun2023test})
                         & 0.455 & 0.572 & 0.611 & 0.723
                         & 0.332 & 0.401 & 0.408 & 0.459
                         & 0.392 & 0.423 & 0.471 & 0.552
                         & 0.233 & 0.303 & 0.359 & 0.452 \\
    TEMPO (Zero Shot)~(\citeyear{tsf_tempo_cao2024tempo})
                        & 0.400 & 0.426 & 0.441 & 0.443
                         & 0.301 & 0.355 & 0.379 & 0.409
                         & 0.438 & 0.461 & 0.515 & 0.591
                         & 0.185 & 0.243 & 0.309 & 0.386 \\
    TimesFM (Zero Shot)~(\citeyear{tsf_timesfm_das2024decoderonlyfoundationmodeltimeseries})
                        & 0.421 & 0.472 & 0.510 & 0.514
                         & 0.326 & 0.399 & 0.434 & 0.451
                         & 0.357 & 0.411 & 0.441 & 0.507
                         & 0.205 & 0.294 & 0.367 & 0.473 \\
    LLM4TS~(\citeyear{tsf_llm4ts_chang2024aligningpretrainedllms})
                    & 0.371 & 0.403 & 0.420 & \underline{0.422}
                    & 0.269 & 0.328 & 0.353 & \textbf{0.383}
                         & \underline{0.285} & \textbf{0.324} & \underline{0.353} & 0.408
                         & 0.165 & 0.220 & \textbf{0.268} & \textbf{0.350} \\
    MOMENT~(\citeyear{tsf_fm_moment_goswami2024momentfamilyopentimeseries})
                        & 0.387 & 0.410 & 0.422 & 0.454
                         & 0.288 & 0.349 & \underline{0.369} & 0.403
                         & 0.293 & \underline{0.326} & \textbf{0.352} & 0.405
                         & 0.170 & 0.227 & 0.275 & 0.363 \\
    \rowcolor{warmgreen!20}
    TTM$_A$*~(\citeyear{tsf_ttms_ekambaram2024neurips})
                          & 0.359 & \colorwr{0.389} & \colorwr{0.405} & 0.448
                          & 0.264 & 0.321 & \underline{0.351} & 0.395
                          & 0.318 & 0.354 & 0.376 & \underline{0.398}
                          & 0.169 & 0.223 & 0.276 & \colorwr{0.342} \\
    \rowcolor{warmred!20}
    TTM$_A$~(\textit{reimpl. from} \cite{tsf_critique_xu2025specialized})
                        & 0.363 & \underline{0.392} & 0.413 & 0.442
                         & \underline{0.262} & \underline{0.324} & 0.351 & 0.392
                         & \textbf{0.283} & 0.332 & \underline{0.353} & \textbf{0.393}
                         & \textbf{0.158} & \textbf{0.213} & \underline{0.269} & 0.369 \\

    \rowcolor{warmgreen!20}
    TTM$_B$*~(\citeyear{tsf_ttms_ekambaram2024neurips})
                          & 0.364 & \colorwr{0.386} & \colorwr{0.404} & 0.424
                          & 0.277 & 0.334 & 0.362 & 0.408
                          & 0.322 & 0.376 & 0.407 & 0.439
                          & 0.171 & 0.238 & 0.304 & 0.410 \\
    \rowcolor{warmred!20}
    TTM$_B$~(\textit{reimpl. from} \cite{tsf_critique_xu2025specialized})
                        & 0.360 & \underline{0.392} & \textbf{0.401} & 0.436
                         & 0.269 & 0.336 & 0.359 & 0.390
                         & 0.291 & \underline{0.325} & 0.363 & 0.419
                         & 0.164 & 0.219 & 0.277 & \textbf{0.350} \\

    TTM$_E$~(\citeyear{tsf_ttms_ekambaram2024neurips})
          & 0.363 & 0.393 & 0.406 & 0.452
          & 0.271 & \textbf{0.324} & 0.357 & \underline{0.388}
          & 0.327 & 0.377 & 0.395 & 0.419
          & 0.178 & 0.238 & 0.290 & 0.379 \\
    CALF~(\citeyear{tsf_aaai25_liu2025calf})
                    & 0.369 & 0.427 & 0.456 & 0.479
                         & 0.279 & 0.353 & 0.362 & 0.404
                         & 0.323 & 0.374 & 0.409 & 0.477
                         & 0.178 & 0.242 & 0.307 & 0.397 \\

    \midrule

    \rowcolor{warmgreen!20}
    DLinear*~(\citeyear{tsf_critique_zeng2023transformers})
                                & 0.375 & 0.405 & 0.439 & 0.472
                                & 0.289 & 0.383 & 0.448 & 0.605
                                & 0.299 & 0.335 & 0.369 & 0.425
                                & 0.167 & 0.224 & 0.281 & 0.397 \\
    \rowcolor{warmred!20}
    DLinear~(\textit{reimpl. from}~\cite{tsf_wang2025timemixer++})
                                & 0.397 & 0.446 & 0.489 & 0.513
                                & 0.340 & 0.482 & 0.591 & 0.839
                                & 0.346 & 0.382 & 0.415 & 0.473
                                & 0.193 & 0.284 & 0.382 & 0.558 \\
    \rowcolor{warmblue!20}
    DLinear~$\spadesuit$~(\textit{reimpl. by us})
                            & 0.379 & 0.421 & 0.452 & 0.523
                            & 0.291  & 0.409  & 0.532  & 0.743
                            & -     & -     & -     & -
                            & -     & -     & -     & - \\
    DASHA~(\citeyear{tsf_critique_xu2025specialized})
                         & 0.369 & 0.401 & 0.430 & 0.478
                         & 0.284 & 0.377 & 0.396 & 0.745
                         & 0.305 & 0.335 & 0.367 & 0.418
                         & 0.169 & 0.224 & 0.290 & 0.378 \\
    AR~(\citeyear{stats_ar_arma_whittle1951hypothesis}) (with d=0)
                        & \underline{0.358} & \textbf{0.390} & 0.410 & 0.424
                         & 0.271 & 0.334 & 0.361 & 0.395
                         & 0.299 & 0.336 & 0.368 & 0.426
                         & \underline{0.163} & \underline{0.218} & \underline{0.271} & \underline{0.366} \\

    Auto-AR~(\citeyear{tsf_auto-arima_hyndman2008automatic})
                         & \textbf{0.357} & \textbf{0.390} & 0.410 & \underline{0.422}
                         & \underline{0.269} & 0.332 & 0.359 & 0.394
                         & 0.299 & 0.336 & 0.368 & 0.426
                         & \underline{0.163} & \underline{0.218} & \underline{0.271} & \underline{0.367} \\

    \bottomrule
  \end{tabular}
}
\end{table}

\subsubsection{Improper and ad hoc error metrics.}\label{sssec:improper-metric}

\paragraph{~\ding{42}~Inaccurate choice of evaluation metric.}
The choice of evaluation metric in forecasting is deceptively complex~(cf.~\S\ref{app:metrics}).
A significant pitfall is using error measures inappropriately or introducing novel
metrics without justification. Unlike classification accuracy or ROC AUC in
standard ML tasks, there is no single dominant error metric in forecasting –
more than 40 different accuracy measures exist, each with its own advantages and
drawbacks.
Additionally, \colorwblkb{aggregating errors} across multiple time series or across forecast horizons requires careful normalization; yet researchers often na\"ively average metrics across series of vastly different scales~\cite{tsf_aaai25_liu2025calf}, implicitly giving more weight to larger-scale series.

For example, normalizing by the series mean treats percentage errors comparably across series, but trends, level shifts, small values, and zeros can make the resulting weights misleading. MAPE is undefined when an actual value is zero, while sMAPE remains denominator-sensitive, is bounded in ways that can mask error magnitude, and is not symmetric between equal-sized over- and underforecasts~\cite{kolassa2026smape}. These issues are particularly acute for intermittent-demand series.

More fundamentally, the error measure determines the point-forecast functional being assessed: squared error elicits the conditional expectation, whereas absolute error --- and MASE with a forecast-independent scaling denominator --- elicits the conditional median; MAPE targets yet another, weighted functional~\cite{kolassa2020best}. Assessing one point forecast with several incompatible losses can therefore conflate forecasting targets. Researchers should state the target functional and pair it with an appropriate error measure. Reporting MSE and MAE for the same output is defensible when conditional symmetry makes the mean and median sufficiently close, or as an explicitly labelled sensitivity analysis; otherwise, separate mean and median forecasts should be evaluated. Introducing new bespoke error measures is discouraged unless necessary, as it hinders comparability and may hide biases. In short, metrics must be selected with care and their known limitations acknowledged when interpreting results.

\paragraph{~\ding{42}~Self-constructed ``fairness'' of comparison.}
The notion of experiment design choices based on ensuring fairness is often, if not always, well-intended with valid, sound reasoning. However, even such well-intended constructions may inadvertently induce a conflict-of-interest if the said fairness design principle introduced in a work benefits or bolsters the appeal of the work itself (e.g. superior results). Specific examples in the context of our discussion include the reporting of N-BEATS~\cite{tsf_nbeats_oreshkin2019n} M4 competition~\cite{makridakis2020m4} results by deconstructing the ensemble approach and reproducing its results based on a single model~\cite{tsf_timesnet} for comparison. If we agree to this `fairness' principle, then question arises whether the TTM~\cite{tsf_ttms_ekambaram2024ttms} models --- dominating the coalesced Table~\ref{tab:ltsf_ett_mse} of ETT datasets results --- should be included as a baseline; since it too utilizes a mixture or ensemble strategy at its core. If not, then where should we place it in? Or, should we instead try to reproduce its result using a less performant version (like disallowing mixing perhaps) like \textsc{TTM}$_{fair}$?!

\subsubsection{Omitting or cherry-picking benchmarks and baseline methods.}\label{sssec:baselines}

Perhaps the most consequential flaw in many TSF benchmarks is the lack of \emph{adequate baseline comparisons}. This can vastly exaggerate the perceived advantage of the new approach. We illustrate this using Table~\ref{tab:ltsf_ett_mse} that coalesces results of AI/ML models evaluating on the LTSF benchmark datasets (Table~\ref{tab:tsf-dataset_statistics}).

Dataset selection creates a parallel source of bias. Roque et al.~\cite{roque2025cherrypicking} show that model rankings in TSF can change substantially when evaluation is restricted to a small, potentially unrepresentative subset of datasets: among the methods they studied, selecting only four datasets could make 46\% appear best-in-class and 77\% appear in the top three. This cherry-picking effect is distinct from our generalist-versus-specialist aggregation argument, but it reinforces the broader point that benchmark composition and aggregation can manufacture an appearance of superiority.

\paragraph{~\ding{42}~Exclusion of na\"ive, relevant statistical baseline methods. }
For instance, Bergmeir et al. recount that a series of recent deep learning papers on long-horizon forecasting (e.g. predicting 2 years of daily exchange rates) benchmarked exclusively against other deep networks, never against a na\"ive baseline – which in fact outperformed all those advanced models~\cite{tsf_critique_hewamalage2023forecast}. Without sanity-check baselines, purported advances may merely be illusions compared to trivial forecasts. Robust benchmarking necessitates comparing new methods against appropriate baseline predictors -- beyond just prior SOTA models—selected according to data and task characteristics. This means employing classical models (ARIMA, ETS) and relevant na\"ive yardsticks (e.g., seasonal na\"ive), all implemented fairly (e.g., correct seasonal periods, tuned ARIMA) to avoid biasing results.

\paragraph{~\ding{42}~Exclusion of Domain-Specific SOTA Baselines and Relevant Benchmarks.}
The issue of inadequate baselining extends beyond just classical or na\"ive methods to the omission of established, domain-specific SOTA models and more challenging, task-relevant benchmark datasets. A stark illustration is found in claims of weather forecasting proficiency based solely on the standard LTSF \textsc{Weather} dataset. While many TSF models report strong performance on this particular dataset, the specialized field of Weather Prediction (WP) uses higher-resolution spatiotemporal datasets that represent a much richer range of dynamics, including regime changes and extreme events such as heatwaves and cold spells (e.g., \textsc{WeatherBench}~\cite{rasp2020weatherbench}, which leverages \textsc{ERA5} reanalysis data~\cite{weather_hersbach2020era5, weather_dataset_wu2023global}) and has developed sophisticated machine learning weather prediction (MLWP) models, such as the SOTA GraphCast~\cite{graphcast_lam2022graphcast}. These domain-specific SOTA models represent the true cutting edge in operational weather forecasting. Yet, they are seldom included as baselines in general TSF papers that claim weather forecasting capabilities. This omission is particularly problematic when general TSF models claim long-horizon weather forecasting (e.g., 30 days or more on the LTSF \textsc{Weather} dataset), whereas even advanced operational WP models like GraphCast typically provide reliable forecasts for up to only 10-12 days. Such discrepancies highlight a disconnect, where claims made within the general TSF community might not align with the established capabilities and rigorous standards of the specialized application domain.

\section{Conclusion}\label{sec:conclusion}

This position paper has scrutinized the prevailing evaluation paradigms in AI/ML-based time-series forecasting. We have argued that an over-reliance on benchmarks dominated by strong, persistent periodicities, coupled with insufficient emphasis on robust baselines and task-appropriate metrics, often leads to an illusion of progress. The pursuit of SOTA through marginal gains on such datasets, particularly with computationally expensive and complex deep learning models, may not always translate to genuine methodological advancement or practical utility for diverse real-world forecasting challenges.

To foster more meaningful and reliable progress, we include our constructive take on recent \colorwblkb{alternative views} -- both contrarian and aligned to ours -- in Appendix~\S\ref{sec:alt_views}. Further, we have proposed a set of actionable measures and a call to action for the TSF community, as detailed in Appendix~\S\ref{sec:recommendations}. These center on principles of transparency, rigor, and relevance in evaluation. A summary is reiterated below:

\begin{tcolorbox}[colback=green!5!white, colframe=black!75!green, width=\textwidth, boxsep=5pt, arc=4pt, breakable, title=\textbf{Path to Rigorous TSF Evaluation: Summary}]
    \textbf{Core Recommendations for Evaluation Protocols:}
    \begin{enumerate}[leftmargin=*, labelindent=0pt, itemsep=2.5pt, topsep=2.5pt, parsep=0pt, label=\textbf{\arabic*.}]
        \item \textbf{Task-Specific Evaluation \& Data Relevance:} Explicitly define forecasting tasks (horizon, dimensionality, data characteristics) and utilize benchmarks that faithfully represent these specific scopes to avoid overgeneralization.
        \item \textbf{Standardized, Transparent, \& Reproducible Protocols:} Adopt centralized, community-agreed frameworks (e.g., GIFT-Eval~\cite{tsf_eval_aksu2024gift}) with versioned static splits and, where appropriate, time-stamped live evaluations, alongside open-source code and accessible data.
        \item \textbf{Robust Baselines \& Objective-Aligned Metrics:} Mandate comparisons against na\"{i}ve, classical statistical (e.g., ARIMA, ETS), and simple ML baselines. Select metrics that reflect the practical objectives and decision-making costs of the application.
        \item \textbf{Multimodal, Context-Aware, \& Diverse Benchmarking:} Develop and adopt benchmarks requiring integration of external information and exhibiting a wider spectrum of real-world complexities beyond simple periodicity.
    \end{enumerate}
    \vspace{0.5em}
    \hrule
    \vspace{0.5em}
    \textbf{Call to Action for Stakeholders:}
    \begin{itemize}[leftmargin=*, labelindent=0pt, itemsep=2.5pt, topsep=2.5pt, parsep=0pt, label=\textbf{\ding{72}}]
        \item \textbf{Authors:} Prioritize rigorous evaluation, transparency, and critical assessment of claimed advancements over marginal gains.
        \item \textbf{Reviewers \& Editors:} Mandate higher rigor, scrutinize methodologies thoroughly, and act as gatekeepers against unreliable findings.
        \item \textbf{Research Community:} Foster a culture valuing truthful progress, support collaborative benchmarking, and champion interdisciplinary learning.
    \end{itemize}
\end{tcolorbox}
\vspace{0.5em}

In closing, improving how we benchmark TSF models is vital for translating research into tangible real-world impact. By embracing the principles of rigorous, transparent, and task-relevant evaluation, the TSF community can build a stronger foundation for future innovations, ensuring that ``State-of-the-Art'' genuinely signifies robust and reproducible advancements applicable to the diverse challenges of forecasting. Let us collectively make rigorous benchmarking the norm, and impactful progress the outcome.

\section*{Acknowledgements}
RS is supported by Canada NSERC CGS-D Doctoral Grant.
RS acknowledges that resources used in preparing this research were provided, in part, by the Province of Ontario, the Government of Canada through CIFAR, and companies sponsoring the Vector Institute \href{https://vectorinstitute.ai/partnerships/current-partners/}{https://vectorinstitute.ai/partnerships/current-partners/}.
CB is supported by a María Zambrano Fellowship that is funded by the Spanish Ministry of Universities and Next Generation funds from the European Union. The work is also supported by Grant PID2023-149128NB-I00 funded by MICIU/AEI /10.13039/501100011033 and by ERDF, EU. It is also partially supported by the I+D+i project granted by C-ING-250-UGR23 co-funded by ``Consejería de Universidad, Investigación e Innovación'' and the European Union related to FEDER Andalucía Program 2021-27.

\clearpage\newpage

\bibliographystyle{plainnat}
\bibliography{paper}

\newpage
\appendixpage
\DoToC
\appendix
\clearpage

\section{Alternative Views}\label{sec:alt_views}

While this paper posits a critical stance on current AI/ML TSF evaluation paradigms and the purported superiority of complex models like LLMs under such regimes, it is important to acknowledge alternative perspectives and ongoing debates within the research community.

\paragraph{~\ding{42}~Aligned Views and Corroborating Evidence}
Our critique aligns with a growing body of work that questions the unverified progress in TSF. For instance, recent studies have demonstrated that specialized foundation models, despite their complexity, often struggle to consistently outperform well-tuned supervised baselines, particularly when evaluation is rigorous~\cite{tsf_critique_xu2025specialized}. This resonates with our argument that true advancements must be evidenced by clear, reproducible gains over established, simpler methods, rather than being artifacts of specific benchmark configurations or evaluation protocols. Furthermore, the general pitfalls in forecast evaluation, such as metric misuse, inadequate baselining, and reproducibility challenges, have been extensively documented~\cite{hewamalage2023forecast, tsf_talk_bergmeir2024fundamental}, underscoring the systemic nature of the issues we address. These works reinforce our call for more principled and robust evaluation practices in TSF research.

\paragraph{~\ding{42}~Unaligned and Contrarian Perspectives}
Conversely, some research directions and theoretical arguments present a more optimistic or alternative outlook on the role and capabilities of large-scale models, particularly LLMs, in the TSF domain.

One prominent line of argument, exemplified by recent position papers (e.g., \cite{tsf_jin2024positionlargelanguagemodels}), advocates for the inherent potential of LLMs in TSF, including for complex multivariate series. Proponents suggest that LLMs, pretrained on vast diverse corpora, possess emergent capabilities that can be effectively leveraged for time series analysis and forecasting, often through prompting or lightweight adaptation. For instance, Jin et al.~\cite{tsf_jin2024positionlargelanguagemodels} posit that LLMs \textit{“excel at processing time series tasks.”}
While we acknowledge the potential for LLMs to augment various aspects of the forecasting workflow (e.g., incorporating exogenous textual information, scenario generation), our analysis in \S\ref{sec:llm_tsf_challenges}, supported by contemporary empirical critiques~\cite{tsf_critique_tan2024language, tsf_merrill2024language}, highlights significant weaknesses of current LLMs in core temporal reasoning and forecasting tasks. We maintain that \textbf{claims} of LLM superiority in these core tasks, particularly on established benchmarks, \textbf{necessitate more robust empirical grounding}. This includes consistent, statistically significant improvements over well-established statistical and simpler machine learning baselines, under transparent and reproducible evaluation settings, which, as we argue, is often lacking in current literature.
advocating for such models

A cautious stance is warranted: while LLMs may offer new avenues, their current efficacy for direct, high-accuracy forecasting, especially in outperforming simpler, efficient models on standard benchmarks, remains an open question requiring more rigorous validation.

Another perspective, offered by researchers such as Andrew Gordon Wilson and collaborators~\cite{tsf_llmtime_gruver2024large}, posits that LLMs can be viewed as powerful, large-scale lossy compression models. From this viewpoint, their ability to learn complex patterns and generate coherent sequences might extend to time series data, potentially allowing them to capture underlying dynamics and offer strong zero-shot or few-shot forecasting capabilities. The argument sometimes extends to suggest that traditional ``no free lunch'' theorems might be less applicable in an era where Transformer-based architectures dominate due to their scalability and expressiveness. This theoretical framing is compelling and offers an alternative lens through which to understand LLM behavior. However, our paper's focus remains on the empirical realities of current TSF evaluation. Even if LLMs act as effective lossy compressors, the practical implications for forecasting accuracy on diverse TSF benchmarks must be demonstrably superior. The ``smoothing'' inherent in some compression schemes might be detrimental for series requiring the prediction of sharp discontinuities or high-frequency components, common in many real-world applications.
Furthermore, the empirical challenges associated with Transformer architectures for TSF, such as handling permutation invariance and temporal ordering (as discussed in \S\ref{sec:llm_tsf_challenges}), persist regardless of their theoretical underpinnings as universal approximators or compressors. Thus, while such theoretical views are valuable for guiding future research, they do not obviate the need for the stringent, comparative, and task-relevant empirical evaluation that forms the cornerstone of our critique. The ultimate utility of any model, irrespective of its theoretical elegance, must be borne out by its performance under fair and rigorous scrutiny against appropriate and challenging benchmarks.

In summary, while diverse viewpoints enrich the scientific discourse, the central tenet of this paper—that rigorous, transparent, and appropriately baselined evaluation is paramount for genuine progress in TSF—remains critical regardless of the specific architectural or theoretical paradigm being explored.

\section{Recommendations and Call to Action}
\label{sec:recommendations}

The pursuit of genuine progress in Time-Series Forecasting (TSF) necessitates a critical re-evaluation of our benchmarking and evaluation methodologies. While the community has begun to recognize the importance of robust evaluation, evidenced by initiatives like TFB~\cite{tsf_tfb_qiu2024tfb} and GIFT-Eval~\cite{tsf_eval_aksu2024gift} (which subsumes established benchmarks like the Monash repository~\cite{tsf_dataset_godahewa2021monash} and promotes live, standardized assessment), the adoption of such principled practices is not yet universal. The persistence of ``cascading errors,'' where flawed initial findings from highly-cited works (e.g., \cite{tsf_timellm_iclr24_jin2023time, tsf_gpt4ts_neurips23_zhou2023one}) are built upon by subsequent research (e.g., \cite{tsf_aaai25_liu2025calf}), underscores the urgent need for change. This paper aims to catalyze this change by offering concrete recommendations and a call to action for all stakeholders in the TSF research ecosystem.

\subsection{Key Recommendations}
To address the shortcomings highlighted in \S\ref{sec:critique} and to foster a culture of scientific rigor, we propose the following interrelated recommendations, drawing upon established best practices~\cite{tsf_critique_hewamalage2023forecast} and aligning with recent calls for reform~\cite{tsf_critique_xu2025specialized}:

\paragraph{\ding{42} Delineate Forecasting Task Types and Employ Task-Relevant Data.}
\colorwblkb{TSF is not monolithic}; tasks vary widely in horizon (short vs. long-term), dimensionality (univariate vs. multivariate), output (point vs. probabilistic), and data characteristics (stationary vs. non-stationary, noisy vs. clean). New methods must be evaluated on datasets and settings that faithfully represent their intended use-case. Researchers should explicitly define their task's scope (e.g., ``monthly univariate forecasting of macroeconomic indicators'' or ``high-frequency multivariate electricity load forecasting'') and select benchmarks accordingly. This task-specific delineation, analogous to NLP's distinct benchmarks for translation versus question answering, will prevent inappropriate generalizations and enable more meaningful comparisons. A model excelling at one task type should not be presumed effective for all.

\paragraph{\ding{42} Adopt Standardized, Transparent, and Reproducible Evaluation Protocols.}
The field urgently requires centralized, community-agreed evaluation frameworks. We advocate for live benchmarks with fixed, chronologically sound public train/validation/test splits, automated result verification, and dynamic leaderboards. Initiatives like GIFT-Eval~\cite{tsf_eval_aksu2024gift} exemplify this transparent approach, mitigating selective reporting and ensuring fair comparisons. Crucially, every published TSF result must be accompanied by open-source code and readily accessible data (or clear instructions for obtaining proprietary data), facilitating full reproducibility. Standardized evaluation harnesses and libraries should be leveraged to streamline this process and reduce the likelihood of implementation errors.

\paragraph{\ding{42} Mandate Robust Baselines and Align Evaluation with Forecasting Objectives.}
New TSF methods must demonstrate clear superiority over relevant and strong baselines. This includes: (a) na\"{i}ve forecasts (e.g., last value, seasonal na\"{i}ve); (b) well-tuned classical statistical models (e.g., ARIMA, ETS); and (c) simple yet effective ML models (e.g., linear models like DLinear~\cite{tsf_critique_zeng2023transformers}, lightweight neural networks). The choice of baselines must be appropriate for the data's characteristics (e.g., seasonal na\"{i}ve for clearly seasonal series). Furthermore, the target functional and evaluation metric must reflect the practical objective of the forecast. For example, MSE assesses a conditional-mean forecast, whereas MAE assesses a conditional-median forecast; probabilistic forecasts require proper scores such as CRPS. Multiple metrics may be reported as sensitivity analyses, but each should be paired with the point forecast it elicits, unless an explicit assumption such as conditional symmetry makes the targets effectively coincide~\cite{kolassa2020best}.

\paragraph{\ding{42} Embrace Multimodality, Contextual Information, and Real-World Complexity.}
Many current TSF models, especially generic large-scale architectures, struggle with incorporating external context or reasoning about multimodal data, limiting their applicability to complex real-world scenarios. Future benchmarks should actively promote the development of models that can leverage such ancillary information. This includes tasks requiring the integration of textual reports with time series (e.g., news headlines for economic forecasting, weather reports for energy demand), image data, or other categorical/event-based signals. As argued by Bergmeir (2024)~\cite{tsf_talk_bergmeir2024fundamental}, true breakthroughs are more likely to arise from integrating domain knowledge and diverse data modalities than from simply scaling up univariate models. Benchmarks that reward the effective use of rich context will drive innovation towards more causal, interpretable, and practically useful forecasting systems.

\paragraph{\ding{42} Complement Static Benchmarks with Living Evaluation.}
Frozen test sets remain valuable for controlled reproducibility, but repeated adaptation to public benchmarks and possible training-data contamination can inflate apparent generalization. Impermanent~\cite{garza2026impermanent} provides a complementary design: it evaluates forecasts sequentially on continuously updated GitHub activity streams, including issues, pull requests, pushes, and new stargazers from widely followed repositories. Because outcomes become available only after forecasts are submitted, such living benchmarks can measure temporal robustness under genuine distributional change. They should complement rather than replace static benchmarks, with time-stamped submissions, delayed scoring, versioned protocols, and archived snapshots preserving auditability and reproducibility.

Adherence to these interrelated recommendations will ensure that new TSF models are rigorously evaluated on appropriate tasks, against strong baselines, using meaningful metrics, and within transparent frameworks, ultimately fostering true scientific progress.

\subsection{Call to Action}

The current state of benchmarking in time-series forecasting research demands urgent improvement. We issue a collective call to action to the AI/ML community engaged in TSF: let us collaboratively elevate the standards of our evaluations. Specifically, we urge authors, reviewers, and conference organizers to embrace and implement the following:

\vspace{1ex}\noindent \textbf{\ding{72} For Authors:} Prioritize \emph{rigorous evaluation} over incremental performance claims. Adhere to the best practices outlined: employ multiple, relevant datasets; include a comprehensive suite of na\"{i}ve, classical, and simple ML baselines; scrupulously avoid data leakage; and report results using appropriate metrics, including statistical significance where applicable. Ensure transparency by releasing code, data splits, and detailed experimental setups. Before claiming state-of-the-art, critically assess if improvements are robust across diverse, fair benchmarks or merely artifacts of a specific experimental design. If your method targets a niche, clearly define its scope and avoid generalizing beyond supported evidence.

\vspace{1ex}\noindent \textbf{\ding{72} For Reviewers and Editors:} Mandate \emph{higher rigor} in submitted TSF studies. Scrutinize evaluation methodologies as thoroughly as model architectures. Reject papers lacking adequate baseline comparisons, employing questionable metrics, or failing to address reproducibility. Encourage ablation studies that test the robustness of findings (e.g., sensitivity to data splits, performance against simpler alternatives). Resist being swayed by marginal gains on narrow benchmarks if fundamental evaluation principles are neglected. By upholding strong evaluation standards, the review process becomes a crucial gatekeeper against the proliferation of unreliable or irreproducible findings.

\vspace{1ex}\noindent \textbf{\ding{72} For the Research Community at Large:} Foster a culture that values \emph{truthful, verifiable progress} over superficial leaderboard chasing. This involves applauding studies that provide thorough, critical evaluations, even if they reveal that simpler methods are competitive or superior. Organize workshops, challenges, and competitions with meticulous, transparent designs—perhaps reviving the spirit of the M-competitions within the ML context—to benchmark methods fairly. Support and expand collaborative benchmarking platforms (e.g., GIFT-Eval~\cite{tsf_eval_aksu2024gift}). Champion interdisciplinary collaboration, enabling the ML community to learn from established forecasting principles while contributing innovative, scalable solutions. By bridging communities and adhering to rigorous standards, we can avoid relearning old lessons and focus on genuine advancements.

\medskip
Improving how we benchmark TSF models is not merely an academic exercise; it is fundamental to translating research into real-world impact. Forecasts underpin critical decisions in diverse sectors. Flawed evaluations risk the deployment of suboptimal models, with potentially significant negative consequences. The path forward, however, is clear. By addressing known pitfalls and embracing comprehensive, rigorous benchmarking as outlined in this paper (and by others, e.g., \cite{tsf_talk_bergmeir2024fundamental}), we can substantially enhance the reliability and practical value of TSF research. We urge the community to embrace this call to action—to scrutinize our evaluation methods as closely as our models. The outcome will be a more robust foundation for future innovation and a discipline that can confidently substantiate its claims of progress. Let us make rigorous benchmarking the norm, and genuine, impactful progress the ultimate outcome.

\section{Miscellaneous and Details}\label{app:sec:misc}
\subsection{The Seeking SOTA Experiment}\label{app:ssec:seeking-sota}
Here we design and propose a thought experiment to aid intuitive understanding and illustration of our arguments to follow. In this experiment, we posit one of the authors as a candidate for the ``best overall human athlete'' on the planet, based on performance across a diverse set of downstream athletic tasks.
We empirically demonstrate, through a constructed scenario, that this author can achieve a superior weighted average score, effectively a form of SOTA performance in this aggregated evaluation, beating eight global top performers in their respective specializations.
This outcome is achieved despite the author not being the single best performer in any individual task, while maintaining a consistently high percentile ranking (typically $>$90th percentile) across each task.

\begin{table}[htbp]
\centering
\caption{Human evaluation results across diverse downstream tasks. The \colorwgb{known} best performances for specialists (world records or top ranks) are in \colorwgb{green and bold}. Hypothetical performances for specialists in non-primary disciplines are obtained by averaging over ChatGPT-4o~\cite{openai2023gpt4} and Gemini-2.5 Pro's~\cite{anil2023gemini} estimate (presented as plain text). The Author's true performances are in \colorwg{green}. The \textbf{best value} in each metric column is in \textbf{bold}. The global ranking percentile (estimated for a relevant population) is shown in parentheses adjacent to performance values.
}
\label{tab:thought_experiment}
\renewcommand{\arraystretch}{1.3}
\resizebox{\textwidth}{!}{
    \begin{tabular}{@{}l c c c c c c c c c@{}}
    \toprule
    & \multicolumn{3}{c}{Running Times}
            & \multicolumn{2}{c}{Cricket (T20I)}
                & \multicolumn{2}{c}{Soccer (CF)}
                    & Chess & Weighted \\
        \cmidrule(lr){2-4} \cmidrule(lr){5-6} \cmidrule(lr){7-8} \cmidrule(lr){9-9} \cmidrule(lr){10-10}
    Athlete & 100m (s) & 200m (s) & 10K (m:s) & Bat. Avg.  & Bowl. Avg. & Goals/G & Assists/G & Elo & Avg. \\
    \midrule
    Usain Bolt (JAM)          & \colorwgb{9.58 (100\%)} & \colorwgb{19.19 (100\%)} & 38:00 (85\%)
                                    & 5 (10\%) & 150 (5\%)
                                    & 0.10 (50\%) & 0.05 (40\%)
                                    & 800 (10\%)
                                    & 0.93 \\
    Joshua Cheptegei (UGA)    & 11.50 (80\%) & 23.00 (75\%) & \colorwgb{26:24 (100\%)}
                                    & 3 (5\%) & 180 (2\%)
                                    & 0.02 (20\%) & 0.01 (15\%)
                                    & 700 (5\%)
                                    & 0.92 \\
    Virat Kohli (IND)         & 11.80 (75\%) & 24.00 (70\%) & 45:00 (70\%)
                                    & \colorwgb{51.75 (99.9\%)} & 90.00 (20\%)
                                    & 0.05 (40\%) & 0.02 (30\%)
                                    & 1100 (30\%)
                                    & 0.91 \\
    Rashid Khan (AFG)         & 12.50 (65\%) & 25.50 (60\%) & 50:00 (60\%)
                                & 12.96 (40\%) & \colorwgb{14.50 (99.9\%)}
                                & 0.01 (15\%) & 0.01 (10\%)
                                & 900 (15\%)
                                & 0.90 \\
    Lionel Messi (ARG)        & 11.20 (85\%) & 23.50 (70\%) & 42:00 (78\%)
                                & 2 (2\%) & 200 (1\%)
                                & \colorwgb{0.79 (100\%)}
                                & \colorwgb{0.37 (100\%)}
                                & 1000 (20\%) & 0.94 \\
    Cristiano Ronaldo (POR)   & 10.90 (90\%) & 22.50 (80\%) & 40:00 (80\%)
                                & 1 (1\%) & 250 (0.5\%)
                                & \colorwgb{0.72 (99\%)} & \colorwgb{0.22 (98\%)}
                                & 950 (18\%)
                                & 0.92 \\
    Magnus Carlsen (NOR)      & 14.00 (60\%) & 28.50 (50\%) & 55:00 (50\%)
                                & 0.5 (0.5\%) & 300 (0.1\%)
                                & 0.08 (45\%) & 0.04 (35\%)
                                & \colorwgb{2830 (100\%)} & 0.96 \\
    Ding Liren (CHN)          & 15.00 (40\%) & 31.82 (30\%) & 65:00 (30\%)
                                    & 0.2 (0.2\%) & 400 (0.05\%)
                                    & 0.005 (5\%) & 0.002 (2\%)
                                    & \colorwgb{2734 (99.9\%)} & 0.95 \\
    \midrule
    \textbf{Author}        & \colorwg{12.05 (91\%)} & \colorwg{26.90 (90\%)}  &  \colorwg{40:54 (96\%)}
                                & \colorwg{38.00 (91\%)} & \colorwg{20.00 (90\%)}
                                &  \colorwgb{0.60 (96\%)} & \colorwgb{0.70 (100\%)}
                                & \colorwg{2290 (99\%)}
                                & \textbf{0.98} \\
    \bottomrule
    \end{tabular}
}
\end{table}

The results in Table~\ref{tab:thought_experiment} illustrate a scenario where the Author, despite not holding a world-record or being the top specialist in any single discipline, achieves the highest ``Weighted Avg.'' score. This is facilitated by consistently strong (e.g., >90th percentile) performances across all four diverse categories of sports and a hypothetical evaluation scheme that rewards such generalist consistency. For instance, the Author's soccer Assists/Game (A/G) of 0.70 is assigned a 100th percentile ranking, showcasing how metric scaling or contextual interpretation within an evaluation framework can significantly influence outcomes. While specialists like Usain Bolt or Magnus Carlsen are undisputed champions in their domains (100th percentile), their performance drops considerably in non-specialist areas. Conversely, the Author's broad competence translates to a superior aggregate score under this specific, multi-task evaluation regime.
This contrived example serves as an intuitive analogy for the core arguments we present regarding the evaluation of global TSF models.

\section{Evaluation Metrics for Time-Series Forecasting}
\label{app:metrics}

\subsection{Common Evaluation Metrics}

Table~\ref{tab:tsf-metric-formulas} presents metrics that are frequently used to evaluate time-series forecasting (TSF) models across classical statistical approaches, modern deep learning systems, and large-scale competitions~\cite{hyndman2008forecasting, makridakis2018statistical, athanasopoulos2011tourism}:

\begin{table}[h]
    \centering
    \caption{Common evaluation metrics used in time-series forecasting.}
    \label{tab:tsf-metric-formulas}
    \renewcommand{\arraystretch}{1.6}
    \resizebox{0.99\textwidth}{!}{
    \begin{tabular}{l l}
        \toprule
        \textbf{Metric} & \textbf{Formula} \\
        \midrule
        \textbf{Mean Squared Error (MSE)} &
        \( \displaystyle \text{MSE} = \frac{1}{H} \sum_{i=1}^{H} \left( y_{T+i} - \hat{y}_{T+i} \right)^2 \) \\[2pt]

        \textbf{Mean Absolute Error (MAE)} &
        \( \displaystyle \text{MAE} = \frac{1}{H} \sum_{i=1}^{H} \left| y_{T+i} - \hat{y}_{T+i} \right| \) \\[2pt]

        \textbf{Mean Absolute Percentage Error (MAPE)} &
        \( \displaystyle \text{MAPE} = \frac{100}{H} \sum_{i=1}^{H} \frac{|y_{T+i} - \hat{y}_{T+i}|}{|y_{T+i}|} \) \\[2pt]

        \textbf{Symmetric Mean Absolute Percentage Error (sMAPE)} &
        \( \displaystyle \text{sMAPE} = \frac{200}{H} \sum_{i=1}^{H} \frac{|y_{T+i} - \hat{y}_{T+i}|}{|y_{T+i}| + |\hat{y}_{T+i}|} \) \\[14pt]

        \textbf{Mean Absolute Scaled Error (MASE)} &
        \( \displaystyle \text{MASE} = \frac{1}{H} \sum_{i=1}^{H} \frac{|y_{T+i} - \hat{y}_{T+i}|}{\frac{1}{T-H-m} \sum_{j=1}^{T-m} |y_j - y_{j-m}|} \) \\[16pt]

        \textbf{Overall Weighted Average (OWA)} &
        \( \displaystyle \text{OWA} = \frac{1}{2} \left[ \frac{\text{sMAPE}}{\text{sMAPE}_\text{Na\"{i}ve2}} + \frac{\text{MASE}}{\text{MASE}_\text{Na\"{i}ve2}} \right] \) \\
        \bottomrule
    \end{tabular}
    }
\end{table}

\paragraph{Explanation.}
In the above, \( y_{T+i} \) is the ground-truth value, and \( \hat{y}_{T+i} \) is the model forecast at horizon \( i \), for \( i = 1, \dots, H \). The variable \( H \) is the forecasting horizon, and \( m \) denotes the seasonal periodicity (e.g., \( m=12 \) for monthly data with yearly seasonality).

\textbf{MSE} and \textbf{MAE} are commonly reported in AI/ML literature due to their simplicity and compatibility with gradient-based optimization. They assess different point-forecast targets: MSE elicits the conditional expectation, while MAE is proportional to the pinball loss for the 50\% quantile and elicits the conditional median~\cite{kolassa2020best}. Both are scale-dependent and are not directly comparable across series without an explicit aggregation or normalization scheme.

\textbf{MAPE} and \textbf{sMAPE} express error in relative percentage terms and are widely used in forecasting competitions. MAPE is undefined when an actual is zero and can strongly overweight small actuals. sMAPE is bounded but still requires an explicit convention when both forecast and actual are zero, treats equal-sized over- and underforecasts differently, and can behave pathologically with small denominators~\cite{kolassa2026smape}.

\textbf{MASE} scales absolute forecast error by the in-sample error of a na\"ive seasonal forecast, enabling scale-free comparison across series~\cite{hyndman2021forecasting}. Because its numerator is an absolute loss, it retains the conditional-median target and can favor zero forecasts for intermittent, low-volume count data.

\textbf{OWA}, introduced in the M4 forecasting competition~\cite{makridakis2018statistical}, combines sMAPE and MASE into a single score normalized relative to the seasonal na\"ive (Na\"ive2) method. An OWA score of 1.0 corresponds to matching Na\"ive2, but the composite does not by itself identify a single conventional point-forecast functional.

\subsection{Limitations of MSE and MAE in Long-Horizon Forecasting Benchmarks}
\label{app:limitations-mse-mae}

\begin{table}[h]
    \centering
    \caption{Comparison of common time-series forecasting evaluation metrics.
    The \emph{elicited functional} maps the predictive distribution to the
    (possibly non-singleton) set of expected-loss
    minimizers~\cite{kolassa2020best}; for strictly positive outcomes the
    $|Y|^{-1}$-weighted median coincides with the classical $(-1)$-median, and
    MAPE is undefined when $Y=0$. \emph{Scale invariant} refers to
    multiplicative rescaling $y \mapsto cy$, $c>0$ (not to shifts).
    \emph{Sensitivity to extreme errors} and \emph{reported form} describe
    loss behaviour and output units and do not rank the metrics.
    \emph{Lead-time aware} refers to the metric's construction; any metric can
    be reported per-horizon. MAPE is undefined when $Y=0$, while sMAPE requires a convention when $Y=\hat{Y}=0$.}
    \label{tab:tsf-metric-properties}
    \vspace{0.5em}
    \renewcommand{\arraystretch}{1.3}
    \resizebox{\textwidth}{!}{
    \begin{tabular}{lcccccc}
        \toprule
        \textbf{Metric}
        & \thead{Elicited\\Functional}
        & \thead{Scale\\Invariant}
        & \thead{Sensitivity to\\Extreme Errors}
        & \thead{Reported\\Form / Units}
        & \thead{Built-in Baseline\\Normalization}
        & \thead{Lead-time\\Aware} \\
        \midrule
        \textbf{MSE}   & cond.\ mean                 & \xmark & Quadratic                       & squared units                    & \xmark & \xmark \\
        \textbf{MAE}   & cond.\ median               & \xmark & Linear                          & original units                   & \xmark & \xmark \\
        \textbf{MAPE}  & $|Y|^{-1}$-weighted median  & \cmark & Linear; $|Y|^{-1}$-weighted     & percentage                       & \xmark & \xmark \\
        \textbf{sMAPE} & nonstandard minimizer                 & \cmark & Bounded; saturates at $200\%$   & bounded percentage $[0,200\%]$   & \xmark & \xmark \\
        \textbf{MASE}  & cond.\ median (fixed scale) & \cmark & Linear (scaled)                 & ratio to in-sample na\"ive scale & \cmark & \xmark \\
        \textbf{OWA}   & nonstandard (composite)     & \cmark & Component-dependent             & index (Na\"ive2 $=1$)            & \cmark & \xmark \\
        \bottomrule
    \end{tabular}
    }
\end{table}
\vspace{1em}


While \textbf{MSE} and \textbf{MAE} are the dominant metrics in machine learning-based time-series forecasting (TSF), particularly in deep learning research, they exhibit several important limitations when used in isolation—especially on the long-horizon TSF (LTSF) benchmarks such as ETT, ECL, Traffic, and Weather (Table~\ref{tab:tsf-dataset_statistics}).

First, both MSE and MAE are \colorwblk{scale-dependent}, making them sensitive to the magnitude of the target variable. In multivariate datasets such as Traffic (862 channels) or Electricity (321 channels), different series may operate on vastly different scales (e.g., urban vs. rural sensors), causing the aggregated error to be dominated by high-magnitude series. As a result, improvements on low-variance or low-volume components can be masked entirely. Scale-normalized metrics can facilitate cross-series comparison, but they do not automatically make it fair: their denominators alter the implicit weighting of series and must be justified for the task.

Second, squared error elicits the \colorwblk{conditional expectation}, which is strongly influenced by extreme outcomes~\cite{kolassa2020best}. Its quadratic form is appropriate when the expectation is the intended forecast or when large errors carry commensurately high decision costs (e.g., in safety-
critical or financial risk domains); it is not an intrinsic defect of MSE. The relevant question is whether the benchmark intends to evaluate a conditional-mean forecast. If another functional is required, the forecast and scoring rule should change together rather than interpreting MSE's sensitivity to large deviations in isolation. For instance, a model that makes a \underline{single large mistake on one time step} may receive a higher MSE than a model that consistently underperforms but avoids large spikes. This is particularly problematic in datasets like Exchange-Rate or Illness, which exhibit irregular or intermittent dynamics.

Third, both MSE and MAE \colorwblk{lack sensitivity to \textit{temporal structure} or \textit{forecasting horizon}}. In long-horizon settings (e.g., predicting 96 or 192 time steps ahead), the aggregation of errors over the full horizon can obscure how well a model performs at different lead times. Two models with similar MSE may differ substantially in their error trajectories --- e.g., one performs well in the short term and deteriorates quickly, while another maintains stable performance across the full horizon. Without horizon-stratified metrics or error curves, these distinctions remain hidden.

Fourth, these metrics \colorwblk{do not themselves compare performance with a seasonal or other na\"ive baseline}, which matters for LTSF benchmarks such as ETTm2 and Weather. A seasonal na\"ive model (e.g., repeating past day/hour values) can achieve surprisingly low MSE on such datasets. Explicitly reporting performance relative to an appropriate na\"ive forecast can reveal whether a model improves on periodic extrapolation; any scaling or composite score used for that comparison must still be justified for the target functional.

No metric suite is universally appropriate for LTSF. Evaluation should first declare the target functional or decision objective, then use a consistent forecast and scoring rule. Multiple metrics can be informative as explicitly labelled sensitivity analyses, but applying incompatible losses to one undifferentiated point forecast can obscure rather than clarify performance. Cross-series studies should additionally disclose their aggregation weights, report horizon-stratified results, and compare against relevant na\"ive baselines.

\subsection{Non-stationarities in Time Series Data}\label{app:ssec:non-stationarities}

\begin{table}[htb]
    \centering
    \captionsetup{skip=5pt}
    \caption{Non-Stationarity based task delineation in time series and corresponding statistical methodologies}
    \label{tab:ts_nonstationarity}
    \renewcommand{\arraystretch}{1.2}
    \resizebox{\textwidth}{!}{

        \begin{tabular}{@{}lllp{5.5cm}@{}}
            \toprule
                \makecell{\textbf{Non-} \\ \textbf{Stationarities}} &

                \makecell{\textbf{Statistical} \\ \textbf{Variants}} &
                \makecell{\textbf{Example} \\ \textbf{Problem Domains}} &
                \makecell{\textbf{Statistical Methodologies} \\ (e.g., Naive Baselines) } \\
            \midrule
            Trend & Mean & Global temperature rise & Detrending and differencing (ARIMA, ETS with trend) \\
            Seasonal & Mean/Covariance & Quarterly sales, electricity demand
                                                & Seasonal adjustment (SARIMA, STL decomposition) \\

            Structural Breaks & Mean/Variance & Economic shocks, policy changes
                                                & Regime-switching models (Markov Switching AR, piecewise models) \\

            Volatility & Variance & Financial returns & Time-varying volatility models (GARCH, EGARCH) \\

            Distributional & Distribution shape & Concept evolution in user behavior & Non-parametric/adaptive modeling (kernel density, quantile regression) \\

            Concept Drift & Data relationships & User-item interaction in streaming & Online/continual learning (dynamic Bayesian models, online forests) \\

            Cyclostationarity & Periodic statistics & EEG, vibration signals    & Spectral/time-frequency methods (wavelet transform, cyclostationary models) \\
            \bottomrule
        \end{tabular}
    }
\end{table}

\section{Details of Long Horizon Time-series Datasets} \label{app:sec:ltsf_datasets}

\begin{table}[htb]
\centering

\caption{Summary of Time Series Forecasting Datasets.}\label{tab:tsf-dataset}
\resizebox{\textwidth}{!}{
    \begin{tabular}{@{}lccccc@{}}
    \toprule
    Datasets    & ETTh1 \& ETTh2    & ETTm1 \&ETTm2     & ECL       & Weather   & Traffic   \\ \midrule
    Channels    & 7                 & 7         & 322       & 21        & 862         \\
    Frequency   & hourly            & 15-min    & hourly    & hourly    & hourly        \\
    Timesteps   & 17,420            & 69,680         & 26,304         & 52,704         & 17,544       \\
    Train/Val/Test Split & 12/4/4 months & 12/4/4 months & 15/3/4 months & 28/10/10 months & 7:1: \\ \bottomrule
    \end{tabular}
}
\vspace{-5pt}
\end{table}

\subsection{Electricity Transformer Temperature --- \textsc{ETT}}
\label{app:ssec:ltsf_ett}

The Electricity Transformer Temperature (ETT) datasets~\cite{tsf_zhou2021informer} monitor key indicators of oil-filled power transformers, collected over a two-year period from two separate counties in China. These datasets are provided at two granularities: hourly (ETTh1, ETTh2) and 15-minute intervals (ETTm1, ETTm2). Each data point includes seven features: high-usefulness fuel flow (HUFL), high-usefulness load (HULL), medium-usefulness fuel flow (MUFL), medium-usefulness load (MULL), low-usefulness fuel flow (LUFL), low-usefulness load (LULL), and the target variable, Oil Temperature (`OT'). The standard split for these datasets typically involves 12 months for training, 4 months for validation, and 4 months for testing, resulting in specific \texttt{test\_start\_index} values depending on the granularity (e.g., 11,520 for hourly, 46,080 for 15-minute data, marking the beginning of the test set after the validation period).

\begin{figure*}[ht!]
    \centering
    \begin{subfigure}[b]{\linewidth}
        \centering
        \includegraphics[width=0.98\linewidth]{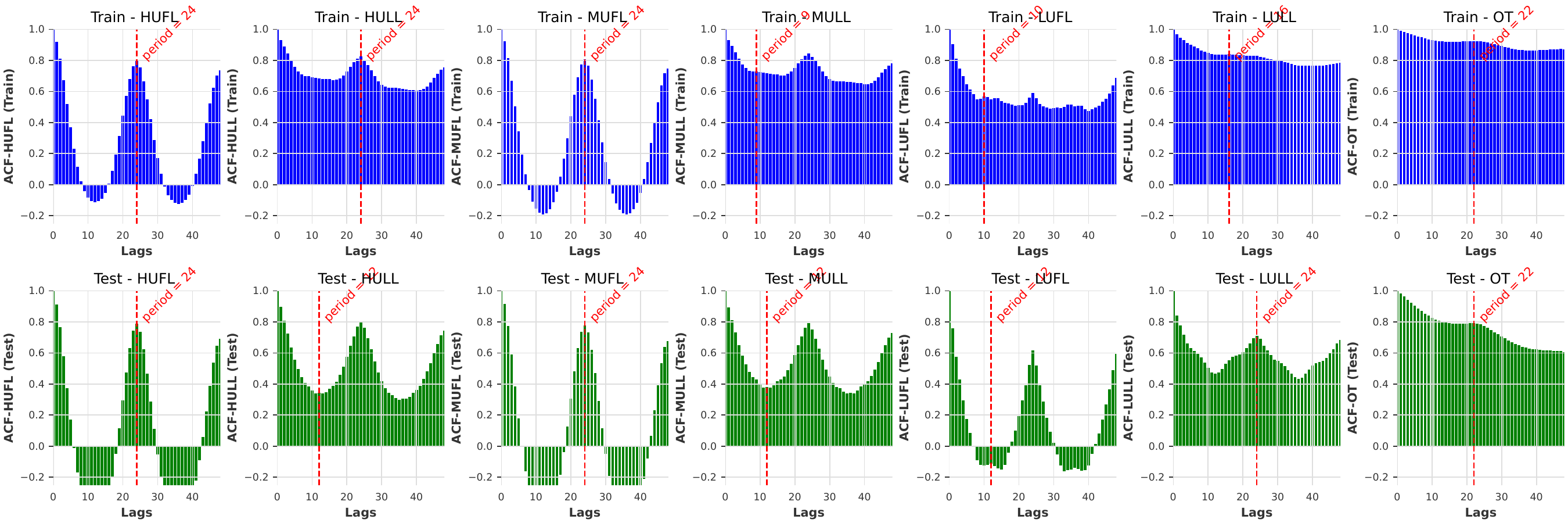}
        \caption{ACF plots for ETTh1 features (Standardized Train vs. Test data).}
        \label{fig:acf_etth1}
    \end{subfigure}
    \vspace{0.5em}
    \begin{subfigure}[b]{0.49\textwidth}
        \centering
        \includegraphics[width=\linewidth]{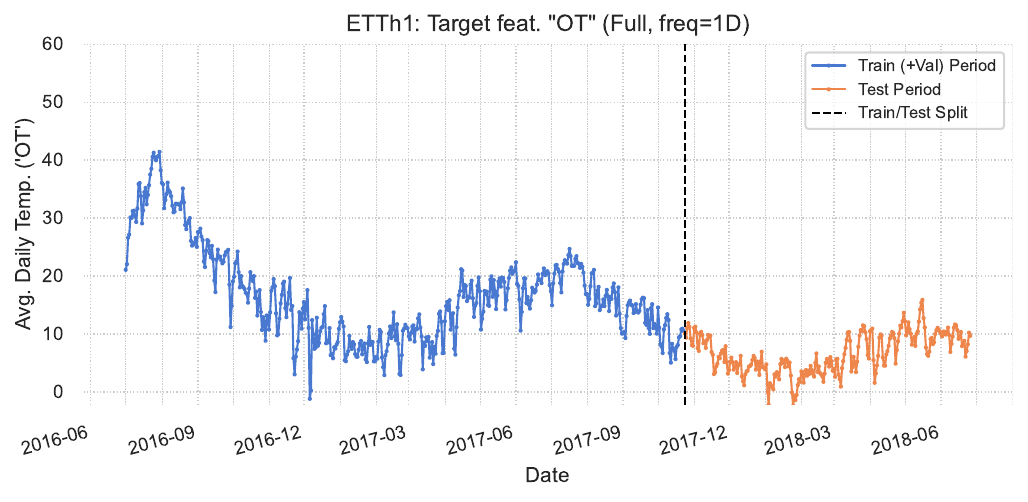}
        \caption{ETTh1 `OT': Full view (daily avg).}
        \label{fig:etth1_full}
    \end{subfigure}
    \hfill
    \begin{subfigure}[b]{0.49\textwidth}
        \centering
        \includegraphics[width=\linewidth]{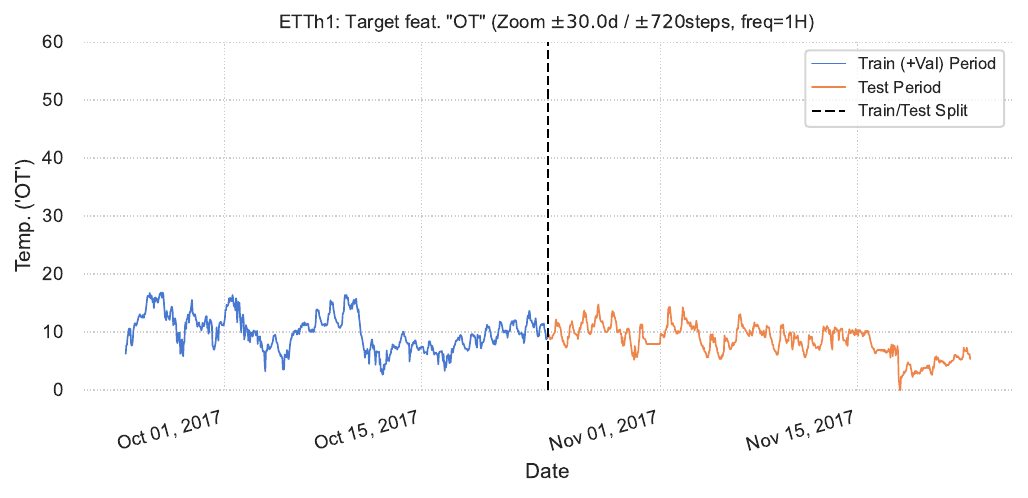}
        \caption{ETTh1 `OT': Zoom ($\pm 30$d, hourly).}
        \label{fig:etth1_zoom}
    \end{subfigure}
    \caption{ETTh1 Dataset: (a) Autocorrelation functions showing persistent daily periodicity (e.g., lag 24) across train/test splits for all features. (b) Full daily average view of the `OT' target. (c) Zoomed hourly view of `OT' around the train/test split, showing continuity of patterns. Target plots share y-max of 60.}
    \label{fig:ett_etth1_analysis}
\end{figure*}

\begin{figure*}[ht!]
    \centering
    \begin{subfigure}[b]{\linewidth}
        \centering
        \includegraphics[width=0.98\linewidth]{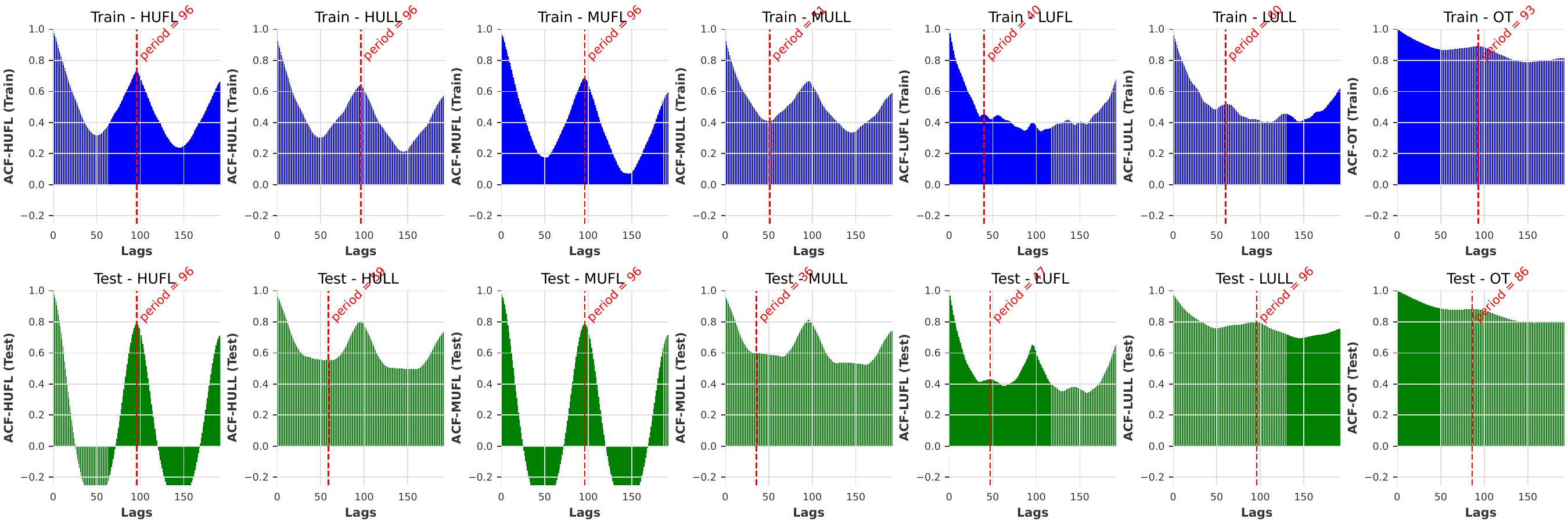}
        \caption{ACF plots for ETTm1 features (Standardized Train vs. Test data).}
        \label{fig:acf_ettm1}
    \end{subfigure}
    \vspace{0.5em}
    \begin{subfigure}[b]{0.49\textwidth}
        \centering
        \includegraphics[width=\linewidth]{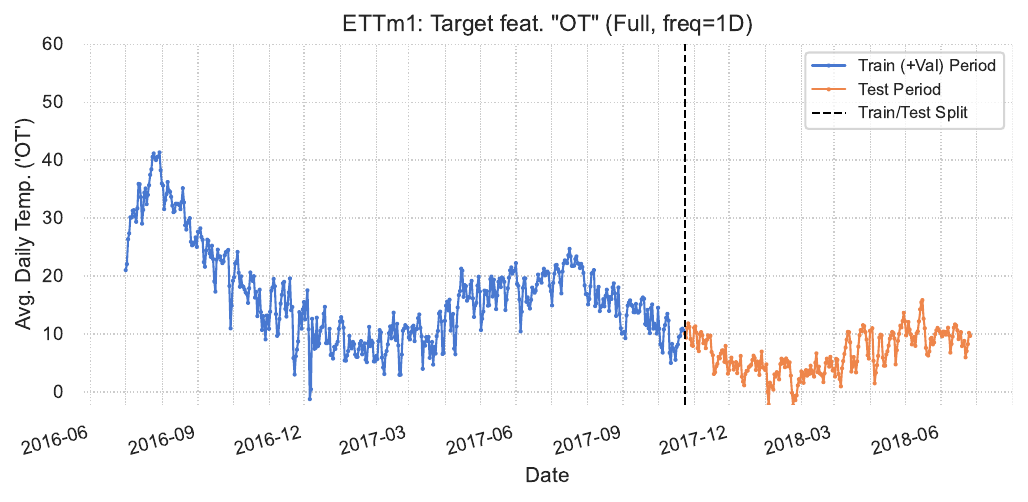}
        \caption{ETTm1 `OT': Full view (daily avg).}
        \label{fig:ettm1_full}
    \end{subfigure}
    \hfill
    \begin{subfigure}[b]{0.49\textwidth}
        \centering
        \includegraphics[width=\linewidth]{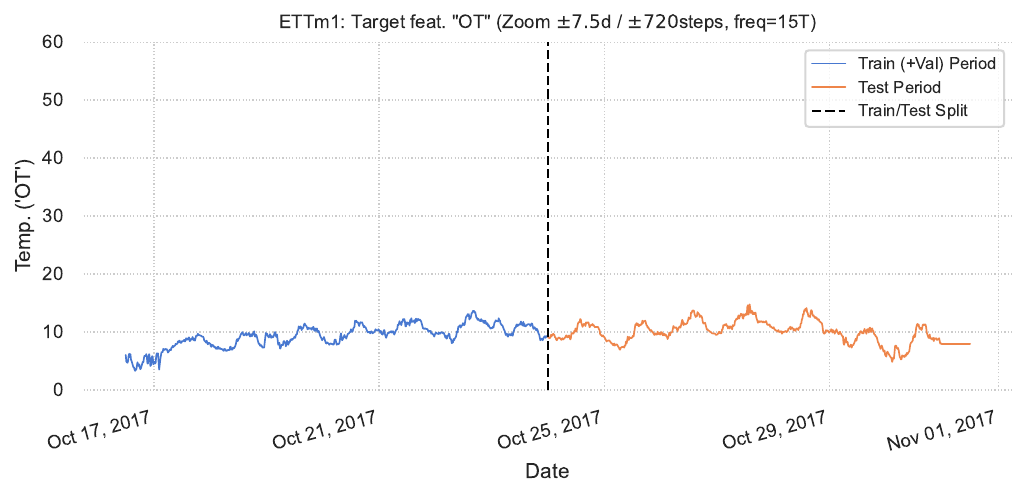}
        \caption{ETTm1 `OT': Zoom ($\pm 7.5$d, 15-min).}
        \label{fig:ettm1_zoom}
    \end{subfigure}
    \caption{ETTm1 Dataset: (a) Autocorrelation functions revealing strong daily periodicity (e.g., lag 96 for 15-min data) persisting across train/test splits. (b) Full daily average view of `OT'. (c) Zoomed 15-minute view of `OT' ($\pm 7.5$ days, i.e., $\pm 720$ steps) around the split. Target plots share y-max of 60.}
    \label{fig:ett_ettm1_analysis}
\end{figure*}

\begin{figure*}[ht!]
    \centering
    \begin{subfigure}[b]{\linewidth}
        \centering
        \includegraphics[width=0.98\linewidth]{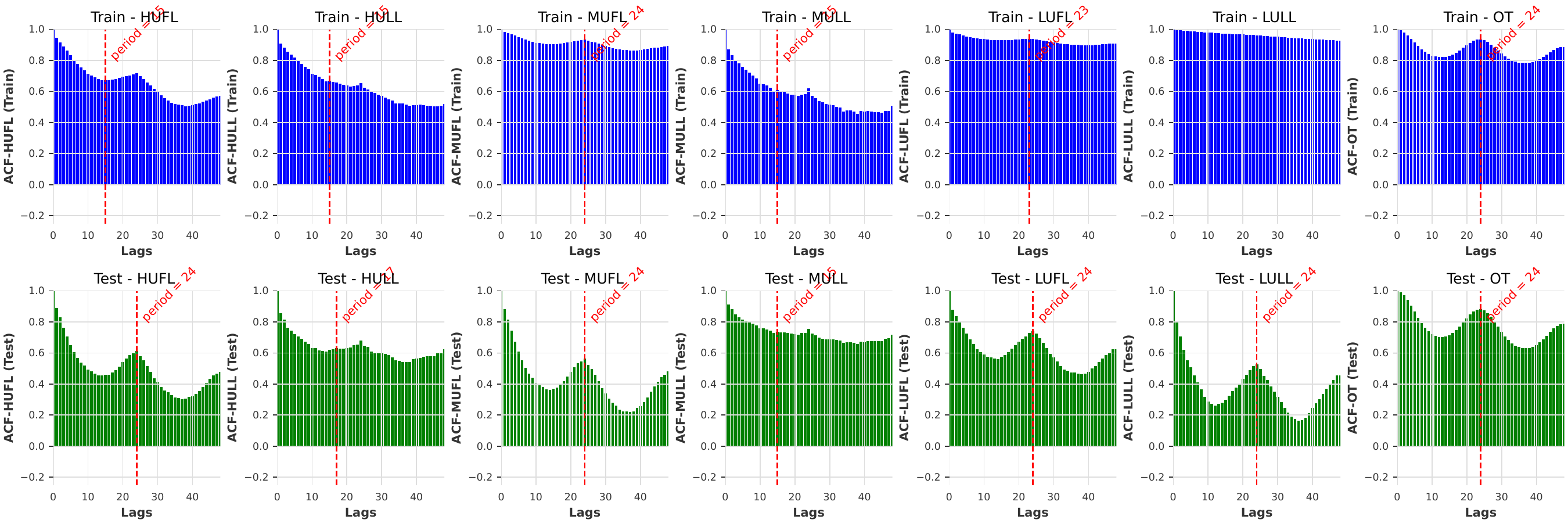}
        \caption{ACF plots for ETTh2 features (Standardized Train vs. Test data).}
        \label{fig:acf_etth2}
    \end{subfigure}
    \vspace{0.5em}
    \begin{subfigure}[b]{0.49\textwidth}
        \centering
        \includegraphics[width=\linewidth]{ett/etth2_target_OT_etth2_full_daily_for_subfigure.pdf}
        \caption{ETTh2 `OT': Full view (daily avg).}
        \label{fig:etth2_full}
    \end{subfigure}
    \hfill
    \begin{subfigure}[b]{0.49\textwidth}
        \centering
        \includegraphics[width=\linewidth]{ett/etth2_target_OT_etth2_zoom_pm720steps_origfreq_for_subfigure.pdf}
        \caption{ETTh2 `OT': Zoom ($\pm 30$d, hourly).}
        \label{fig:etth2_zoom}
    \end{subfigure}
    \caption{ETTh2 Dataset: (a) Persistent daily periodicity evident in ACF plots across train/test splits. (b) Full daily average view of `OT'. (c) Zoomed hourly view of `OT' at the train/test boundary. Target plots share y-max of 60.}
    \label{fig:ett_etth2_analysis}
\end{figure*}

\begin{figure*}[ht!]
    \centering
    \begin{subfigure}[b]{\linewidth}
        \centering
        \includegraphics[width=0.98\linewidth]{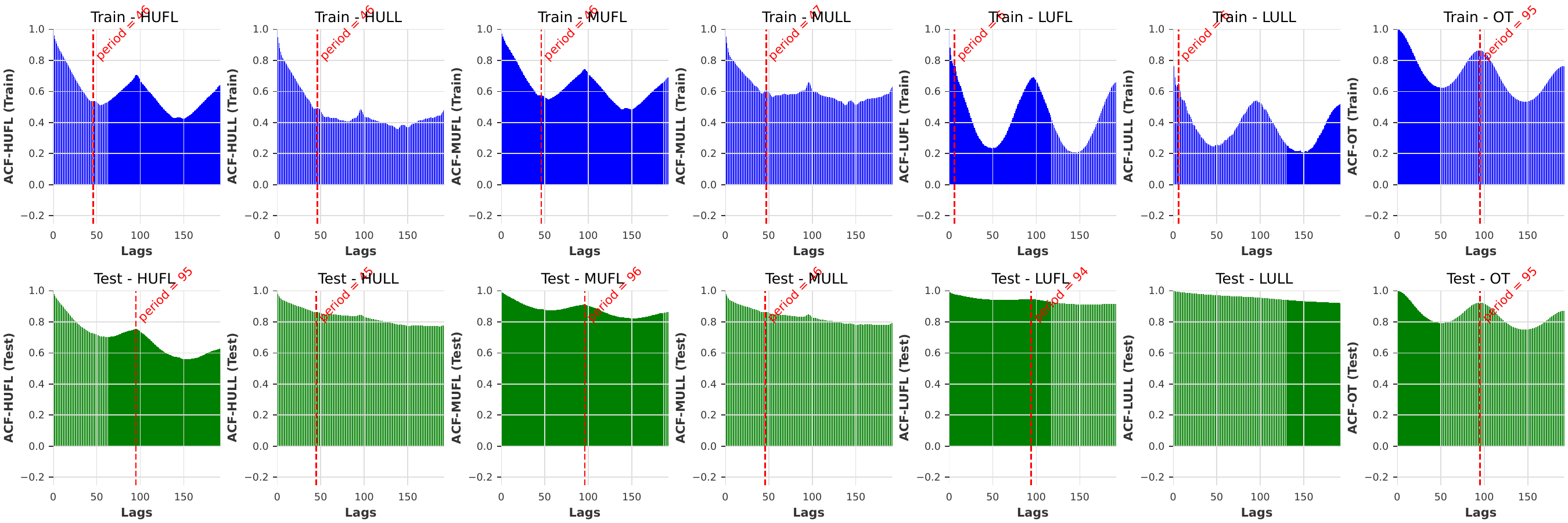}
        \caption{ACF plots for ETTm2 features (Standardized Train vs. Test data).}
        \label{fig:acf_ettm2}
    \end{subfigure}
    \vspace{0.5em}
    \begin{subfigure}[b]{0.49\textwidth}
        \centering
        \includegraphics[width=\linewidth]{ett/ettm2_target_OT_ettm2_full_daily_for_subfigure.pdf}
        \caption{ETTm2 `OT': Full view (daily avg).}
        \label{fig:ettm2_full}
    \end{subfigure}
    \hfill
    \begin{subfigure}[b]{0.49\textwidth}
        \centering
        \includegraphics[width=\linewidth]{ett/ettm2_target_OT_ettm2_zoom_pm720steps_origfreq_for_subfigure.pdf}
        \caption{ETTm2 `OT': Zoom ($\pm 7.5$d, 15-min).}
        \label{fig:ettm2_zoom}
    \end{subfigure}
    \caption{ETTm2 Dataset: (a) Strong daily periodicity (e.g., lag 96) shown in ACF plots, consistent across train/test phases. (b) Full daily average of `OT'. (c) Zoomed 15-minute view of `OT' ($\pm 7.5$ days) around the split. Target plots share y-max of 60.}
    \label{fig:ett_ettm2_analysis}
\end{figure*}

A critical characteristic of the ETT datasets is their strong and persistent periodicity, evident across multiple features and both train and test splits. Figure~\ref{fig:acf_etth1}, \ref{fig:acf_ettm1}, \ref{fig:acf_etth2}, and \ref{fig:acf_ettm2} illustrate the Autocorrelation Function (ACF) plots for all seven features (standardized) for the ETTh1, ETTm1, ETTh2, and ETTm2 datasets, respectively, comparing the ACF structure in the training data (first 60-80\% of data before the test split) versus the test data.

The ACF plots consistently reveal dominant periods, often around lag 24 for hourly data (ETTh1, ETTh2) corresponding to daily cycles, and around lag $24 \times 4 = 96$ for 15-minute data (ETTm1, ETTm2), also reflecting daily patterns. The striking similarity in ACF patterns and identified main periods between the train and test segments for most features underscores the stationary nature of these periodicities. This inherent, stable periodicity implies that forecasting models, even simpler linear ones like LTSF-Linear~\cite{tsf_critique_zeng2023transformers}, can achieve strong performance by effectively capturing these recurring cycles, questioning the necessity for overly complex architectures if these patterns are the primary drivers of predictability in these benchmarks.

Alongside the ACF analysis, time series plots of the `OT' target variable provide further insight for each dataset, showing a daily resampled full view and a zoomed-in view at the original frequency around the train/test split.

\subsubsection{\textsc{ETT1} (ETTh1 and ETTm1)}

\paragraph{ETTh1 -- Periodicity and Target Visualization}
The hourly ETTh1 dataset exhibits clear daily cycles.
\FloatBarrier

\paragraph{ETTm1 -- Periodicity and Target Visualization}
The 15-minute ETTm1 dataset shows finer-grained daily patterns.
\FloatBarrier

\subsubsection{\textsc{ETT2} (ETTh2 and ETTm2)}

\paragraph{ETTh2 -- Periodicity and Target Visualization}
Similar to ETTh1, ETTh2 displays consistent daily periodic behavior.
\FloatBarrier

\paragraph{ETTm2 -- Periodicity and Target Visualization}
The ETTm2 dataset, with its 15-minute granularity, also shows robust daily cycles.
\FloatBarrier

\clearpage\newpage

\subsection{Electricity Consuming Load --- \textsc{ECL}}
\label{app:ssec:ltsf_ecl}

The Electricity Consuming Load (ECL)~\cite{ECL} dataset tracks the hourly electricity consumption (in kWh) for 321 anonymous clients. The version commonly used in LTSF benchmarks (e.g., from~\cite{tsf_wu2021autoformer}, `\textit{electricity.csv}') comprises 26,304 hourly time steps, spanning approximately 3 years (e.g., from 2016 to 2019, though the exact start/end can vary by preprocessing). The dataset features 322 columns: a `date' column, 320 columns representing the consumption of individual clients (typically labeled '0' through '319' in preprocessed files), and a designated target column, often preprocessed and named `OT' for consistency across benchmarks, which might represent the consumption of a specific client (e.g., client 320 if following a pattern) or an aggregate. For this `OT' target, the observed maximum value is 6035.0 kWh.

The standard procedure for this dataset involves a train/validation/test split, often cited as 15 months for training, 3 months for validation, and 4 months for testing. Similar to other utility datasets like ETT, ECL exhibits strong daily and weekly seasonalities due to human activity patterns, as well as annual seasonality reflecting temperature-driven consumption changes (e.g., heating and cooling). These pronounced periodicities are generally consistent across the train and test splits, making them a significant factor in model performance. Consequently, models that effectively capture these cyclical components, including simpler statistical or linear approaches, can achieve competitive results on this benchmark.

\subsection{\textsc{Traffic}} \label{app:ssec:ltsf_traffic}

The Traffic dataset~\cite{tsf_dataset_traffic} comprises hourly road occupancy rates from 862 sensors on San Francisco Bay Area freeways, collected by the California Department of Transportation (Caltrans) Performance Measurement System (PeMS) over a period of approximately two years (mid-2016 to early 2018 in the standardized version used in AI/ML papers since and from~\cite{tsf_wu2021autoformer}). The occupancy rates are normalized values between 0 and 1. The full dataset contains 17,544 time steps. For our visualization and typical benchmarking splits, the test period commences after approximately 80\% of the data has been observed (adhering to the adopted \texttt{7:1:1} split ratio).

\begin{figure*}[hbt!]
    \centering
    \begin{subfigure}[b]{0.49\textwidth}
        \centering
        \includegraphics[width=\linewidth]{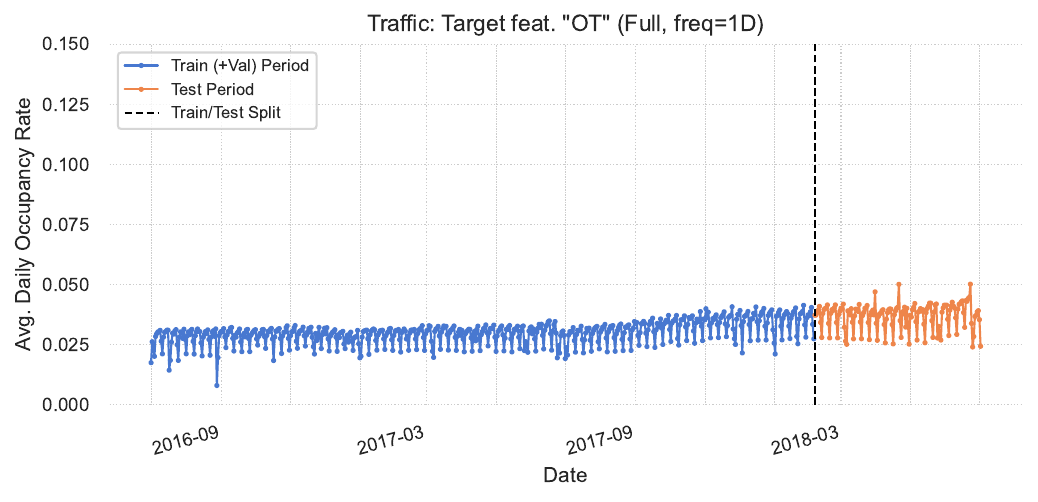}
        \caption{Full daily view (freq=1D).}
        \label{fig:traffic_full_daily}
    \end{subfigure}
    \hfill
    \begin{subfigure}[b]{0.49\textwidth}
        \centering
        \includegraphics[width=\linewidth]{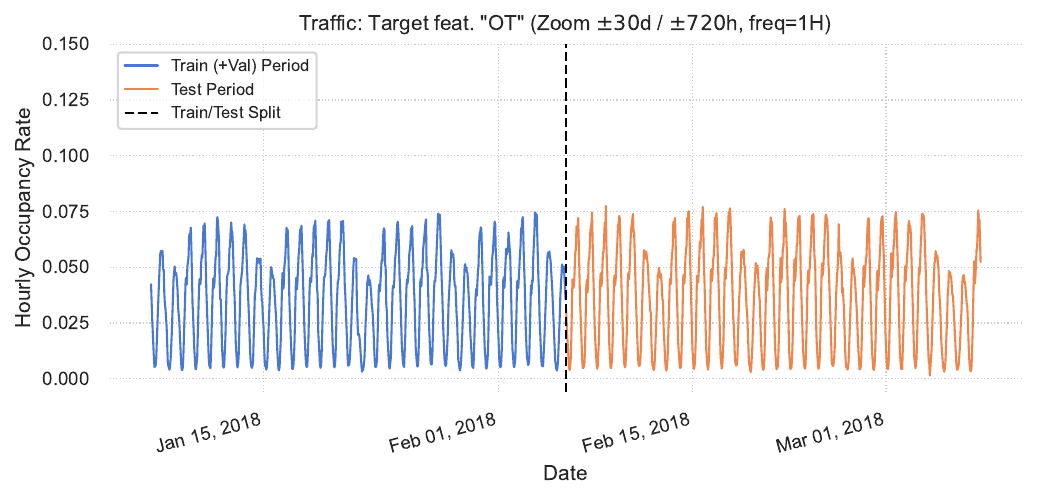}
        \caption{Zoomed hourly view ($\pm 30$ days / $\pm 720$ hours)}
        \label{fig:traffic_zoom_hourly}
    \end{subfigure}
    \caption{Visualization of the Traffic dataset's \textbf{target} feature \textbf{`OT'} (an occupancy rate). (a) The full series is plotted as daily averages to reveal long-term trends and seasonalities. (b) A zoomed-in hourly view around the train/test split highlights the persistence of daily and weekly periodic patterns across the split. Both plots share a consistent y-axis maximum (0.15) for comparable scaling.}
    \label{fig:traffic_visualization}
\end{figure*}

Figure~\ref{fig:traffic_visualization} illustrates the characteristics of a representative target variable, denoted as feature `OT' in the preprocessed version of this dataset. This specific `OT' feature likely represents a chosen sensor's occupancy rate, renamed for consistency across benchmark datasets. The visualization is presented in two panels: a full view of the daily averaged series and a zoomed-in hourly view around the train/test split.

The full daily view (Figure~\ref{fig:traffic_visualization}a) is generated by resampling the original hourly `OT' data to a daily frequency using the mean occupancy rate for each day. This aggregation helps to smooth out high-frequency noise and makes broader patterns, such as weekly seasonality (e.g., differences between weekday and weekend traffic) and any long-term trends, more discernible. We observe a relatively stable baseline occupancy with some fluctuations and a slight upward trend towards the end of the training period, which appears to continue into the test period.

The zoomed-in hourly view (Figure~\ref{fig:traffic_visualization}b) focuses on a window of $\pm 30$ days (equivalent to $\pm 720$ hours) around the designated train/test split point. This detailed perspective clearly reveals strong daily and weekly periodicities inherent in traffic data. The cyclical patterns of peak and off-peak hour traffic, as well as potential differences between weekday and weekend traffic flows, are evident. Notably, these pronounced periodic patterns appear to seamlessly transition from the training segment into the test segment. This observation suggests that models capable of capturing and extrapolating these dominant periodicities (such as simple linear models with seasonal components or even na\"{i}ve seasonal forecasts) might perform surprisingly well on this dataset, potentially challenging the necessity for highly complex architectures if such patterns are the primary drivers of predictability. The consistent y-axis scaling up to 0.15 across both plots helps in comparing the magnitude of daily averages versus hourly fluctuations.

\subsection{\textsc{ExchangeRate}} \label{app:ssec:ltsf_exchangerate}

The Exchange Rate dataset~\cite{tsf_dataset_exchange_lai2018modeling} comprises daily exchange rates of eight countries: Australia, British Pound, Canada, Switzerland, China, Japan, New Zealand, and Singapore, typically against the US Dollar. The data spans from 1990 to 2016, providing 7,588 daily observations. Due to variations in recording (e.g., missing weekend/holiday data), the time series can exhibit irregularities if treated purely as integer-indexed, though modern benchmarks often use a continuous date index. The version of the dataset used here, sourced from~\cite{tsf_wu2021autoformer}, preprocesses one of these exchange rates (or a derivative) as the target variable, consistently named `OT' across several benchmark files. The test period typically begins after observing 80\% of the historical data.

\begin{figure*}[ht!]
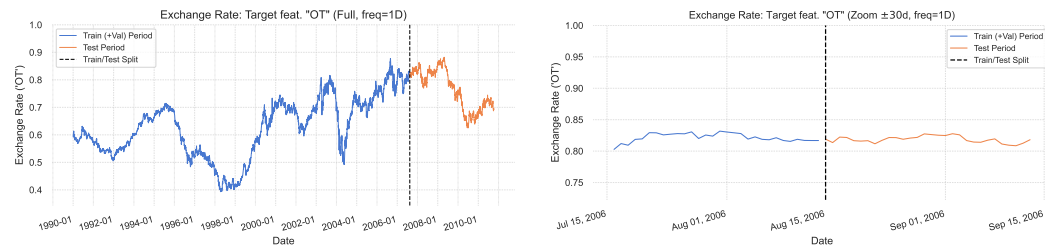

    \centering
    \begin{subfigure}[b]{0.49\textwidth}
        \centering
        \includegraphics[width=\linewidth]{exchange_target_OT_full_daily_ymax_1_0_for_subfigure.pdf}
        \caption{Full time series view of target feature `OT' (freq=1D).}
        \label{fig:exchange_full_daily}
    \end{subfigure}
    \hfill
    \begin{subfigure}[b]{0.49\textwidth}
        \centering
        \includegraphics[width=\linewidth]{exchange_target_OT_zoom_pm30d_ymax_1_0_for_subfigure.pdf}
        \caption{Zoomed daily view ($\pm 30$ days, freq=1D) around the train/test split for target feature `OT'.}
        \label{fig:exchange_zoom_daily}
    \end{subfigure}
    \caption{Visualization of the Exchange Rate dataset's target feature `OT'. (a) The full daily series reveals long-term trends and periods of varying volatility. (b) A zoomed-in daily view of $\pm 30$ days around the train/test split illustrates local dynamics. Both plots utilize a consistent (shared) y-axis maximum of 1.0.}
    \label{fig:exchange_visualization}
\end{figure*}

Figure~\ref{fig:exchange_visualization} illustrates the behavior of this `OT' target variable. Panel (a) shows the full daily time series, while panel (b) provides a zoomed-in view of $\pm 180$ days around the train/test split.

The full daily view (Figure~\ref{fig:exchange_visualization}a) showcases characteristics common to financial time series: periods of relative stability interspersed with higher volatility (volatility clustering), and discernible long-term trends or regime shifts. Unlike datasets with strong, regular seasonality like \textit{traffic} or \textit{electricity consumption}, exchange rates are often modeled as (near) random walks or processes with time-varying parameters, making long-horizon forecasting particularly challenging. The plot shows how the `OT' feature evolves over several years, exhibiting non-stationarity.

The zoomed-in view (Figure~\ref{fig:exchange_visualization}b), while still at a daily frequency, focuses on the $\pm 30$-day interval surrounding the train/test split. This allows for an examination of how short-to-medium term patterns, such as local trends or volatility persistence, behave across this critical juncture.
This closer inspection helps to understand the local dynamics transitioning from the training to the test period. While strong, deterministic periodicities are absent, short-term persistence or momentum might be observable.
The challenge for forecasting models lies in capturing the complex, often non-linear dynamics and adapting to changes in trend and volatility. The apparent continuity of behavior across the split in this zoomed view might suggest that recent historical patterns are somewhat indicative of near-future movements, though the inherent stochasticity remains high. The y-axis is scaled consistently across both plots to facilitate comparison of value ranges.

Lastly, we note that this dataset is not contiguous temporally. The  continuity problem arises with some missing observations or time steps. This is characteristic of financial datasets where non-trading days are absent. For exchanges, further issues may emanate accounting for any timezone differences among the exchange currencies' countries.  Due to these inconsistencies concerning the dates, the resulting preprocessed time series used is integer-indexed.

\subsection{\textsc{Weather}} \label{app:ssec:ltsf_weather}

The Weather~\cite{tsf_dataset_weather_angryk2020multivariate} dataset contains local climatological data for nearly 1,600 U.S. locations, spanning 4 years from 2010 to 2013, with data points collected every 1 hour. Each data point consists of the target value ``wet bulb'' and 11 climate features. The train/val/test split is 28/10/10 months.

We apply a measure periodicity using peak ACF calculation on the train and test data splits of this dataset. Table~\ref{tab:acf_periods_weather} lists the periodicities across features (or, channels/dimensions) in the splits.

\begin{table}[!htbp]
    \centering
    \caption{Peak autocorrelation (ACF) periods for applicable features in the \textsc{Weather}~\cite{tsf_dataset_weather_angryk2020multivariate} dataset.}
    \label{tab:acf_periods_weather}
    \renewcommand{\arraystretch}{1.2}
    \begin{tabular}{l l c c}
        \toprule
        \textbf{Feature} & \textbf{Unit} & \multicolumn{2}{c}{\textbf{Periodicity (ACF Peak)}} \\
        \cmidrule(lr){3-4}
        & & \textbf{Train} & \textbf{Test} \\
        \midrule
        Temperature ($T$) & °C & 141 & 140 \\
        Potential Temperature ($T_{\text{pot}}$) & K & 141 & 140 \\
        Relative Humidity (RH) & \% & 142 & 141 \\
        Saturation Vapor Pressure (VP$_{\text{max}}$) & mbar & 141 & 140 \\
        Vapor Pressure Deficit (VP$_{\text{def}}$) & mbar & 141 & 142 \\
        Air Density ($\rho$) & g/m$^3$ & 139 & 136 \\
        Wind Velocity ($wv$) & m/s & 4 & 102 \\
        Max Wind Velocity (Max $wv$) & m/s & 75 & 92 \\
        Wind Direction ($wd$) & ° & 31 & 4 \\
        Rainfall (Rain) & mm & 15 & 17 \\
        Rain Duration (Raining) & s & 51 & 23 \\
        Shortwave Downward Radiation (SWDR) & W/m$^2$ & 143 & 144 \\
        Photosynthetically Active Radiation (PAR) & mol/m$^2$/s & 143 & 144 \\
        Max PAR (Max PAR) & mol/m$^2$/s & 133 & 143 \\
        Log Temperature ($T_{\text{log}}$) & °C & 142 & 141 \\
        Outdoor Temperature (OT) & °C & 145 & 144 \\
        \bottomrule
    \end{tabular}
    \vspace{0.5em}
    {\centering
        \scriptsize
        \parbox{0.8\linewidth}{
        Weather dataset was originally acquired at \url{https://www.ncei.noaa.gov/data/local-climatological-data/}. We use the standardized version of \texttt{weather.csv} used in AI/ML literature, obtained from~\cite{tsf_wu2021autoformer}.
        }
    \par}
\end{table}

\subsubsection{Weather Datasets in Weather Prediction (WP)}
Weather forecasting/prediction in itself is an active, well-researched domain. Contemporary SOTA models can be dichotomized into numerical weather prediction (NWP) and machine learning weather prediction (MLWP) methodologies. The sub-domain has well defined, rigorous evaluation baselines and publicly available weather datasets like \textsc{WeatherBench}~\cite{rasp2020weatherbench}, \textsc{ERA5}~\cite{weather_dataset_wu2023global, weather_hersbach2020era5}. These weather datasets contain realistic weather phenomena like extreme heat and cold characterized by large anomalies with respect to typical climatology.

\subsubsection{Weather Baselines in Weather Prediction (WP)}
As for SOTA models, the European Centre for Medium-Range Weather Forecasts (ECMWF)’s High Resolution Forecast (HRES) is generally considered the most accurate deterministic NWP-based weather model in the world. Even this top  operational deterministic system in the world produces global 10-day forecasts hourly at 0.1\degC latitude/longitude resolution~\cite{haiden2018evaluation}.
The seminal graph neural network (GNN) based GraphCast~\cite{graphcast_lam2022graphcast}, however, has famously surpassed HRES in computation-commensurate performance --- a feat heralded as another milestone in the success of machine learning (ML)-based approaches. Preceding GraphCast, Pangu-Weather~\cite{weather_mlmodel_panguweather_corringham2019atmospheric} is another strong ML baseline.

\subsection{\textsc{Illness}(ILI)} \label{app:ssec:ltsf_ili}

The national Influenza-like Illness (ILI) dataset, often sourced from CDC data and benchmarked in works like~\cite{tsf_dataset_ili_poorawala2022novel}, tracks \colorwblk{weekly occurrences} of ILI.

The version used here (from~\cite{tsf_wu2021autoformer}, `\texttt{national\_illness.csv}') contains \textbf{966 weekly observations} spanning multiple years (e.g., starting from 2002).
The dataset includes features such as the date, `\% WEIGHTED ILI', `\%UNWEIGHTED ILI', counts for age groups (`AGE 0-4', `AGE 5-24'), total ILI counts (`ILITOTAL'), number of reporting providers (`NUM. OF PROVIDERS'), and a \textbf{target variable} preprocessed under the name \textbf{`OT'} for consistency with other benchmark datasets.
Given its weekly frequency, the forecast horizons for this dataset in AI/ML literature (e.g.,~\cite{tsf_critique_zeng2023transformers}) are shorter, typically $H \in \{24, 36, 48, 60\}$ weeks, as opposed to the longer hourly-equivalent horizons used for other LTSF datasets. The standard 80\%-20\% split is applied for train/validation and test sets.

\begin{figure*}[ht!]
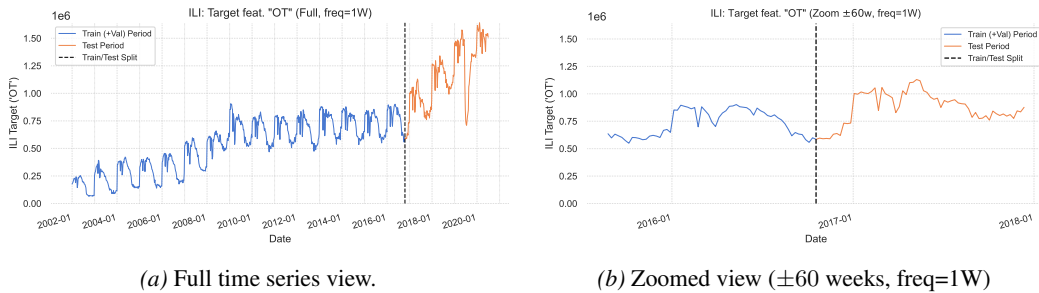

    \centering
    \begin{subfigure}[b]{0.49\textwidth}
        \centering

        \includegraphics[width=\linewidth]{ili_target_OT_full_weekly_for_subfigure.pdf}
        \caption{Full time series view.}
        \label{fig:ili_full_weekly}
    \end{subfigure}
    \hfill
    \begin{subfigure}[b]{0.49\textwidth}
        \centering

        \includegraphics[width=\linewidth]{ili_target_OT_zoom_pm60w_for_subfigure.pdf}
        \caption{Zoomed view ($\pm 60$ weeks, freq=1W)}
        \label{fig:ili_zoom_weekly}
    \end{subfigure}
    \caption{Visualization of the ILI dataset's target feature `OT'.
    (a) The full weekly series clearly shows \colorwblk{strong annual seasonality corresponding to flu seasons.} (b) A zoomed-in weekly view of $\pm 60$ weeks around the train/test split \colorwblk{highlights the consistency of these \textbf{seasonal peaks} and \textbf{troughs} across the split}. Both plots are scaled appropriately from 0 to approximately 1.65 million, reflecting the observed range of the `OT' feature (weekly counts) with minor padding.}
    \label{fig:ili_visualization}
\end{figure*}

Figure~\ref{fig:ili_visualization} provides a visual representation of the `OT' target variable from this dataset. Panel (a) presents the entire weekly time series, and panel (b) shows a detailed view of $\pm 60$ weeks around the train/test split.

\FloatBarrier

The full weekly view (Figure~\ref{fig:ili_visualization}a) prominently displays a strong and regular annual seasonality, with peaks typically occurring during winter months, characteristic of influenza outbreaks in temperate regions. The amplitude of these seasonal peaks can vary from year to year, indicating differing severity of flu seasons.

The zoomed-in perspective (Figure~\ref{fig:ili_visualization}b) on the $\pm 60$-week period surrounding the train/test split further emphasizes the consistent nature of this annual periodicity. The rise and fall of the `OT' values, corresponding to flu seasons, are clearly visible and maintain their pattern across the training and test data boundary. This strong, regular seasonality suggests that models adept at capturing periodic components, including classical decomposition methods or even simpler seasonal baseline models, would likely establish strong performance benchmarks for this dataset. The visual continuity of the pattern across the split implies a high degree of predictability based on these seasonal cycles. The y-axis for both plots is scaled from 0 to an appropriate maximum (e.g., 250,000, based on the observed range of `OT' values) to properly represent the magnitude of the target variable.

\newpage\clearpage

\section{NeurIPS 2025 Position Paper Track: Reviewer Feedback Summary}
\label{app:sec:neurips.2025-reviews}

This paper was submitted to the inaugural NeurIPS 2025 Position Paper Track (Submission \#102) and received a final decision of \textbf{Reject}, despite generally favourable individual reviewer assessments. The rejection reflects the exceptionally high selection threshold applied in this inaugural edition of the track, rather than substantive technical or conceptual deficiencies in the paper. Two of the three reviewers recommended \emph{Accept} or \emph{Weak Accept} (ratings of 7 and 6, respectively), and all three acknowledged the importance and timeliness of the paper's core argument. The Area Chair's borderline-reject meta-review (rating 4) emphasized concerns about novelty relative to existing community discussions and the omission of concurrent large-scale benchmark efforts such as GIFT-Eval.

\subsection*{Summary of Ratings}

\begin{table}[htbp]
\centering
\caption{NeurIPS 2025 Position Paper Track reviewer ratings for Submission \#102. Ratings are on a 1--10 scale; confidence on a 1--5 scale. Sub-dimension scores (Support, Significance, Presentation, Context) are on a 1--4 scale. The mean column includes only the three reviewers and excludes the Area Chair (AC) meta-review.}
\label{tab:neurips2025-ratings}
\renewcommand{\arraystretch}{1.3}
\resizebox{\textwidth}{!}{%
\begin{tabular}{@{}l c c c c c c c@{}}
\toprule
 & \textbf{Reviewer 1} & \textbf{Reviewer 2} & \textbf{Reviewer 3} & \textbf{Area Chair} & \textbf{Mean (reviewers)} \\
\midrule
\textbf{Rating}       & 6 (Weak Accept) & 7 (Accept) & 4 (Borderline Reject) & 4 (Borderline Reject) & \textbf{5.67} \\
\textbf{Agreement}    & 4: agree        & 4: agree   & 3: neither            & 3: neither            & --- \\
\textbf{Confidence}   & 4               & 4          & 2                     & 4                     & \textbf{3.33} \\
\midrule
Support               & 4 (excellent)   & 3 (good)   & 2 (fair)              & ---                   & 3.00 \\
Significance          & 2 (fair)        & 3 (good)   & 2 (fair)              & ---                   & 2.33 \\
Presentation          & 3 (good)        & 3 (good)   & 1 (poor)              & ---                   & 2.33 \\
Context               & 3 (good)        & 3 (good)   & 2 (fair)              & ---                   & 2.67 \\
Discussion            & 3 (possibly)    & 3 (possibly) & 3 (possibly)        & ---                   & 3.00 \\
\bottomrule
\end{tabular}
}
\end{table}

\noindent The mean reviewer rating of \textbf{5.67} (out of 10), computed from the three reviewer scores only (6, 7, and 4), with a mean confidence of \textbf{3.33}, reflects a generally positive but divided reception. Reviewer 1 and Reviewer 2 both recommended acceptance, while Reviewer 3 rated below threshold and focused more strongly on presentation and accessibility. All reviewers confirmed that the submission raises a clear ML-related position and reported no ethical concerns.

\subsection*{Reviewer 1 --- Weak Accept (6/10)}

\paragraph{Summary.} The reviewer correctly identified the two core claims: (1) current TSF benchmarks are unreliable due to dominant periodicity, and (2) future DL-based papers should include classical baselines. The reviewer rated \emph{Support} at 4 (excellent), reflecting strong experimental backing.

\paragraph{Strengths.}
\begin{itemize}
    \item Clearly presents the key observation that reported progress on standard TSF benchmarks may not reflect genuine advances.
    \item Table 3 is highlighted as particularly compelling evidence that DL methods may be learning naive periodic patterns rather than novel dynamics.
    \item Concrete, actionable recommendations for the community (better evaluations, sceptical reviewing, improved benchmarking).
\end{itemize}

\paragraph{Concerns Raised.}
\begin{itemize}
    \item \emph{Known issues}: The reviewer notes that similar concerns had already been raised in prior community discussions.
    \item \emph{Section C.1 typo}: A typo in the appendix was noted.
    \item \emph{General TSF methods}: The reviewer questions whether general-purpose TSF is a sufficiently compelling research direction.
\end{itemize}

\subsection*{Reviewer 2 --- Accept (7/10)}

\paragraph{Summary.} The reviewer praised the paper's clarity, rigour, empirical grounding, and relevance to the NeurIPS community, especially in the context of growing interest in foundation models. All sub-dimension scores were 3 (good); Agreement: 4 (agree).

\paragraph{Strengths.}
\begin{itemize}
    \item The central position---that periodicity-heavy benchmarks overstate DL model performance---is articulated with ``clarity and rigour.''
    \item The \emph{Seeking SOTA} thought experiment is described as a ``compelling'' and illuminating analogy.
    \item The call for taxonomy-specific evaluation protocols and mandatory classical baselines is ``timely and deeply relevant.''
\end{itemize}

\paragraph{Concerns Raised.}
\begin{itemize}
    \item \emph{Scope clarification}: The reviewer recommends stating more explicitly that the critique is aimed primarily at academic benchmark settings.
    \item \emph{Benchmark diversity}: The reviewer suggests broader coverage of hierarchical, multi-resolution, and multimodal datasets.
    \item \emph{Classical method limitations}: The reviewer notes that classical methods have limitations in high-dimensional or exogenous-signal-rich settings.
    \item \emph{Alternative perspectives}: The reviewer highlights reproducibility and LLM-assisted forecasting as relevant topics for broader discussion.
\end{itemize}

\paragraph{Questions Raised.}
The reviewer raised five forward-looking questions concerning hybrid datasets, exogenous covariates, future LLM architectures for temporal reasoning, high-volatility or concept-drift regimes, and shrinkage-based regularization for TSF foundation models.

\subsection*{Reviewer 3 --- Borderline Reject (4/10)}

\paragraph{Summary.} The reviewer acknowledged the paper's good tone in the first three sections and Section 5.1, and agreed that the claim about marginal DL improvements is a meaningful prompt for community reflection. 

\paragraph{Strengths.}
\begin{itemize}
    \item Sections 1--3 and Section 5.1 are praised for setting a good tone.
    \item The core claim about marginal transformer improvements is viewed as a valid and important community prompt.
\end{itemize}

\paragraph{Concerns Raised.}
\begin{itemize}
    \item \emph{Narrative flow (Sections 3--4)}: The transition to the thought experiment is noted as abrupt.
    \item \emph{Accessibility}: The reviewer found parts of the paper less accessible to a broader audience.
    \item \emph{Transformer advantages}: The reviewer asks whether the paper sufficiently discusses settings in which transformers may offer advantages.
\end{itemize}

\subsection*{Area Chair Meta-Review --- Borderline Reject (4/10)}

\paragraph{Summary.} The Area Chair acknowledged the paper's aim and noted that some supporting experiments are present, but raised concert that the paper does not engage with recent large-scale industry benchmark efforts such as GIFT-Eval.

\paragraph{Main Concerns.}
\begin{itemize}
    \item \emph{GIFT-Eval and related benchmark efforts}: The AC highlights the omission of recent large-scale benchmark efforts such as GIFT-Eval.
    \item \emph{Known issues}: The AC notes that several of the paper's concerns are already present in existing community discussions.
\end{itemize}

\subsection*{Overall Assessment}

The paper received a \textbf{final decision of Reject} from the Program Chairs on 26 September 2025 (modified 17 October 2025), in the context of an inaugural position paper track at NeurIPS with an exceptionally high bar for acceptance. The review record shows that two of three reviewers recommended acceptance, both with high confidence (4/5) and agreement (4/5). The paper's position---that TSF evaluation should adopt taxonomy-specific benchmarking and require classical baselines---remains timely and actively debated in the community. We include these reviews as part of the submission record.

\end{document}